\documentclass[lettersize,journal]{IEEEtran}
\usepackage{amsmath,amsfonts}
\usepackage{algorithmic}
\usepackage{algorithm}
\usepackage{array}
\usepackage[caption=false,font=normalsize,labelfont=sf,textfont=sf]{subfig}
\usepackage{textcomp}
\usepackage{stfloats}
\usepackage{url}
\usepackage{verbatim}
\usepackage{graphicx}
\usepackage{cite}
\usepackage{xcolor}
\usepackage{amssymb}
\usepackage{amsthm}
\usepackage{bm}
\usepackage{multirow}
\usepackage{booktabs}
\usepackage{hyperref}
\usepackage{cleveref}

\usepackage{ulem}

\ifdefined\showrevisions
    \newcommand{\revised}[1]{\textcolor{blue}{#1}}
    
    \newcommand{\deleted}[1]{\textcolor{red}{\sout{#1}}}
\else
    \newcommand{\revised}[1]{#1}
    
    \newcommand{\deleted}[1]{} 
\fi

\graphicspath{{figures/}}

\begin{document}

\title{NeuralSSD: A Neural Solver for Signed Distance Surface Reconstruction}

\author{Zi-Chen Xi, Jiahui Huang, Hao-Xiang Chen, Francis Williams, Qun-Ce Xu, Tai-Jiang Mu, Shi-Min Hu
}



\maketitle

\begin{abstract}
We proposed a generalized method, NeuralSSD, for reconstructing a 3D implicit surface from the widely-available point cloud data. 
NeuralSSD is a solver-based on the neural Galerkin method, aimed at reconstructing higher-quality and accurate surfaces from input point clouds. Implicit method is preferred due to its ability to accurately represent shapes and its robustness in handling topological changes. 
However, existing parameterizations of implicit fields lack explicit mechanisms to ensure a tight fit between the surface and input data. To address this, we propose a novel energy equation that balances the reliability of point cloud information. Additionally, we introduce a new convolutional network that learns three-dimensional information to achieve superior optimization results. This approach ensures that the reconstructed surface closely adheres to the raw input points and infers valuable inductive biases from point clouds, resulting in a highly accurate and stable surface reconstruction. 
NeuralSSD is evaluated on a variety of challenging datasets, including the ShapeNet and Matterport datasets, and achieves state-of-the-art results in terms of both surface reconstruction accuracy and generalizability.
\end{abstract}

\section{Introduction}
\IEEEPARstart{S}{urface} reconstruction from point clouds is a fundamental problem in 3D vision and graphics. In practice, point samples are often sparse, noisy, and incomplete due to sensor limitations, occlusions, and acquisition constraints, which makes faithful geometry recovery challenging. Recovering accurate surfaces from such data is critical in robotics, medical imaging, and interactive graphics, where geometric fidelity directly impacts downstream tasks and user experience. This paper targets robust reconstruction under these adverse conditions, aiming to recover watertight, detail-preserving surfaces from real-world point clouds.

\revised{Implicit representations—most commonly signed distance fields (SDFs) and occupancy—are widely adopted for surface reconstruction because they compactly encode complex geometry and naturally accommodate topology changes. In this work we adopt an SDF formulation for robustness and geometric expressiveness, while deferring a concise background on implicit fields and extraction algorithms to Section \ref{sec:related}.}

Existing approaches fall into two families: optimization-based surface fitting and data-driven learning. Classical SDF/variational methods \cite{calakli2011ssd,carr2001reconstruction} reconstruct by interpolating points with hand-crafted regularization to trade off smoothness and detail. These fixed priors are hard to tune, not data-adaptive, and under noisy or incomplete sampling often yield artifacts and limited generalization.
\IEEEpubidadjcol
Learning-based methods either exploit the implicit bias of untrained networks \cite{williams2019deep}—requiring heavy per-scene optimization—or train feed-forward models from large datasets \cite{choy20163d,mescheder2019occupancy}. \revised{Although efficient at inference, they lack mechanisms to guarantee tight adherence to input points and can inherit dataset biases \cite{tatarchenko2019single}, leading to over-smoothing or hallucinated structure.}

\begin{figure*}[!t]
  \centering
  \includegraphics[width=\textwidth]{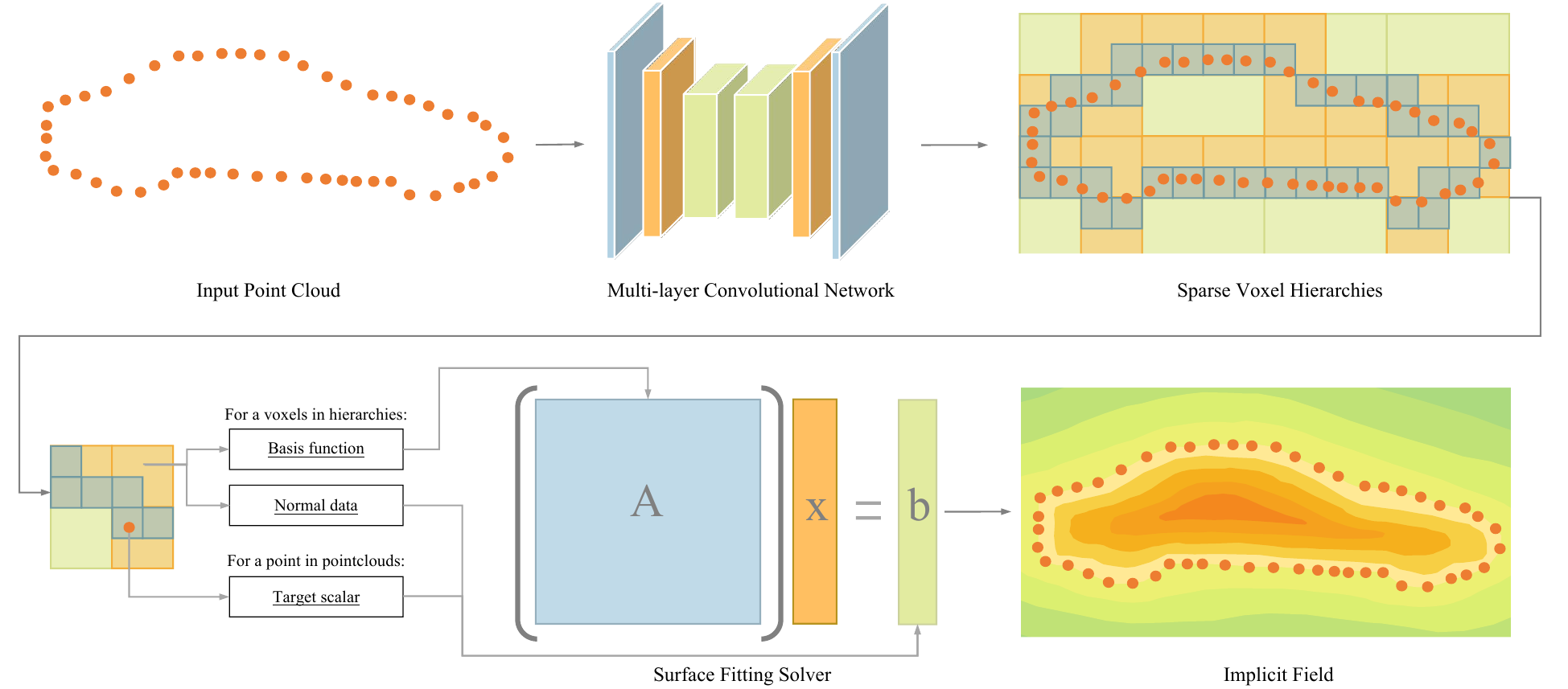}
  \caption{The NeuralSSD pipeline. Given a raw point cloud, we first construct a multi-scale sparse voxel grid. An adaptive sparse network then predicts spatially varying basis functions and normals within this grid. Finally, a closed-form, differentiable SDF solver computes the basis coefficients to reconstruct a watertight surface. This hybrid approach combines a learned, structure-aware prior with a variational optimization that explicitly enforces data fidelity, enabling robust and detailed reconstruction from sparse and noisy inputs.}
  \label{fig:teaser}
\end{figure*}

\revised{To bridge these limitations, we propose NeuralSSD: a unified hybrid framework that couples an adaptive sparse convolutional predictor with a closed-form SDF fitting solver. }The learned component produces a structure-aware sparse voxel scaffold (normals and basis functions), while the optimization component solves a variational fitting problem that explicitly balances data fidelity and learned priors. The solver is fully differentiable, enabling end-to-end training and yielding watertight, noise-robust reconstructions at scale; an overview is shown in Figure \ref{fig:teaser}.

Briefly, NeuralSSD parameterizes the SDF as a spatially varying basis expansion predicted by an adaptive sparse convolutional network over a multi-scale voxel scaffold. Given raw points and inferred normals, a closed-form variational solver computes the basis coefficients with an interpretable regularization that balances data fidelity and learned priors. The pipeline is fully differentiable, enabling end-to-end training and scalable, watertight, detail-preserving reconstruction.

\revised{In summary, the contributions of NeuralSSD can be outlined as follows:
\begin{enumerate}
\item{Unified hybrid framework that enforces point-wise input fidelity while leveraging learned priors, resolving the fidelity--generalization trade-off under sparse/noisy sampling.}
\item{A novel energy formulation that combines normal alignment, Hessian regularization, and point constraints, enabling robust reconstruction from noisy and sparse data.}
\item{Adaptive sparse neural network that infer normals and spatially varying bases on a multi-scale voxel scaffold, scaling to large scenes and producing watertight, detail-rich reconstructions.}
\end{enumerate}
Empirically, NeuralSSD delivers higher accuracy and watertightness under noise and incompleteness, while reducing per-scene optimization time compared to purely optimization-based methods; see Section \ref{sec:exp}.}

\section{Related Works}
\label{sec:related}
This section provides a review of both classical and recent surface reconstruction methods, with a specific focus on point-cloud-based approaches. 
\subsection{Background: Implicit Surface Representations}
\revised{An implicit surface defines a shape as the level set of a scalar field $f$, commonly at threshold $\delta=0$. Formally,}
\begin{equation}
  \mathcal{S} := \{ P \in \mathbb{R}^3 \mid f(P) = \delta \}.
\end{equation}
\noindent where $P\in\mathbb{R}^3$ denotes a 3D point, $f:\mathbb{R}^3\!\to\!\mathbb{R}$ is the scalar field (e.g., SDF), and $\delta\in\mathbb{R}$ is the level-set threshold (commonly $0$).
Two prevalent choices of $f$ are signed distance fields (SDFs), which map each point to its signed distance to the surface (negative inside, positive outside), and occupancy fields, which indicate inside/outside with binary values. When a discrete approximation is required, the implicit surface can be extracted from a volumetric grid using algorithms such as Marching Cubes\cite{lorensen1998marching}.
\cite{curless1996volumetric} is one of the earliest and foundational techniques, utilizes an analytic volumetric running mean fusion strategy to transform scan data into 3D models. To augment the method's robustness against noise and corrupted data, \cite{carr2001reconstruction} incorporates radial basis function interpolates and addresses the reconstruction challenge through low-rank smoothness regularization. \cite{ohtake2005multi} adopt an approach of representing the shape via a combination of multiple parametric surfaces, which are fitted using local geometrical features. Our methodology is influenced by the study by \cite{calakli2011ssd}, the SSD is employed to approximate the signed distance field through the resolution of a variational problem that leverages linear parameterization of function families. Poisson reconstruction \cite{kazhdan2006poisson} and its screened variant \cite{kazhdan2013screened}involve solving partial differential equations (PDEs) to reconstruct surfaces from Hermite data. These techniques have proven to be empirically robust and effective under diverse conditions and have found widespread application in contemporary systems, as exemplified in \cite{vizzo2021poisson}.

\subsection{Learning-Based Surface Reconstruction}
Within the domain of learning-based 3D reconstruction, a myriad of strategies have been developed to optimally leverage data priors for enhanced reconstruction quality, facilitated by the advancement of deep learning technologies and the proliferation of extensive 3D datasets. \cite{chu2021unsupervised} innovatively embedded the inductive bias directly within a 3D CNN architecture, eliminating the need for training data. The NKF framework \cite{williams2022neural} introduced a novel data-dependent kernel, offering improvements over traditional non-learned kernel methods that originate from infinitely wide ReLU networks, as discussed in Neural Splines\cite{williams2021neural},  while NKSR\cite{huang2023neural} also utilizes a data-dependent kernel but limits its spatial extent to boost computational efficiency and employs a gradient-based fitting approach to enhance noise immunity. The concept of transforming 3D points into a feature grid through convolutional architectures was previously suggested in ConvONet\cite{peng2020convolutional} and CIRCLE\cite{chen2022circle} for predicting occupancy fields. POCO\cite{boulch2022poco} enhanced the efficacy and performance of ConvONet by integrating a transformer architecture in place of conventional convolutions. However, both of these methods suffer from prolonged reconstruction times for even modest-sized scenes. In contrast, our feature mapping leverages a hierarchical sparse data structure for heightened efficiency. Sparse data structures have been explored in ASR\cite{ummenhofer2021adaptive} and DOGNN\cite{wang2022dual}, which put forth octree-based convolutional architectures for the reconstruction of expansive scenes. Neural-IMLS~\cite{wang2023neural} directly learns a noise-resistant signed distance function from unoriented raw point clouds in a self-supervised manner. DiGS~\cite{ben2022digs} introduces a divergence-guided constraint to learn a high-fidelity representation from unoriented points, while OG-INR~\cite{koneputugodage2023octree} employs an octree-guided approach to solve the ambiguity of inside/outside labeling for unoriented point clouds. SSRNet\cite{9839681} divided point clouds into different local parts and processes them in parallel.Shape as Points\cite{peng2021shape} employs a method to upsample input points followed by a differentiable Poisson reconstruction, a concept that NGSolver\cite{huang2022neural} expands upon by incorporating learnable basis functions. Point2mesh\cite{hanocka2020point2mesh} demonstrated that mesh deformation over templates can yield high-quality reconstructions by naturally constraining the (self-)priors within the manifold of surfaces. The latest implicit representations, which utilize a singular feed-forward network to parameterize geometry \cite{mescheder2019occupancy}and\cite{ park2019deepsdf}, introduce novel constraints \cite{atzmonsald}and \cite{lipman2021phase} to regularize training. \revised{Other works focus on the learning objective and process. For instance, Neural-Pull~\cite{ma2021neural} introduces a differentiable operation to pull query points towards the surface. To handle noisy inputs, some methods propose to learn from noisy data directly, such as learning a noise-to-noise mapping~\cite{ma2023learning} without clean supervision. ALTO~\cite{wang2023alto} improves reconstruction fidelity and speed by alternating between different latent topologies. InstantNGP~\cite{muller2022instant} leverages a multiresolution hash encoding to accelerate the training and rendering of neural graphics primitives by several orders of magnitude.}

\subsection{Neural Representations for Surfaces}
In the realm of shape parametrization, recent scholarly works have extensively examined the application of neural networks. Various 3D data representations have been employed to enhance representational efficacy. Initial approaches like 3d-r2n2\cite{choy20163d} utilized dense voxel grids. To improve efficiency, subsequent developments introduced sparse and hierarchical data structures such as octrees, as seen in OctNet\cite{riegler2017octnet} and O-CNN~\cite{wang2017cnn}, or other adaptive voxel grids~\cite{liao2018deep} to capture local variations. Vox-Surf\cite{li2022vox} divides the space into sparse voxels that store geometry and appearance information on its corner vertices. \revised{Similarly, patch-based approaches have been proposed to leverage local details. Points2Surf~\cite{erler2020points2surf} learns a prior over local patches combined with global information to improve generalization. DeepLS~\cite{chabra2020deep} stores a grid of latent codes, each representing a local SDF, to enable efficient and detailed reconstruction. } The intricate nature of neural functions necessitates the imposition of regularization strategies to ensure surface smoothness\cite{gropp2020implicit}. Conversely, biases towards certain feature frequencies \cite{sitzmann2020implicit} \cite{tancik2020fourier} or multi-scale representations \cite{martel2021acorn}\cite{takikawa2021neural} have been introduced to enhance the geometric detail in specific areas. BACON~\cite{lindell2022bacon} introduces a network architecture with an analytical Fourier spectrum, allowing for better control over the spectral characteristics of the represented signal. A distinct strand of research addresses implicit fields more systematically: \cite{williams2021neural} presented a methodology to align an infinitely-wide network with point data interpolation via the kernel method, which was subsequently refined into a learnable model in \cite{williams2022neural}. \revised{ImplicitFilter~\cite{li2024implicit} applied an implicit filter to smooth the field while preserving details.} Diverging from direct neural network utilization for shape depiction, our approach employs a spatially-variable set of basis functions and resolves the surface fitting issue through a closed-form solution. Empirical evidence suggests that while our method\'s representational capacity mirrors that of the aforementioned techniques, it facilitates an optimal fit more readily, eliminating the need for labor-intensive optimizations.

\section{Method}
Our method predicts a continuous watertight 3D surface from a point cloud without normal. The method is based on the Neural Galerkin method, which is a variational formulation of the PDE that involves a neural network to infer a set of spatially-varying basis functions that discretize the target implicit field. Given a set of points $\mathcal{P} :=\{ P\in \mathbb{R} ^3\}$, the method constructs a signed distance field on multi-layer sparse voxel hierarchy by a sparse convolutional network, to obtain an expressive surface representation. A mesh can later be constructed from the implicit surface representation using techniques such as the dual marching cubes algorithm.

In the following sections, we will describe the details of our method, and summarize the key steps in Figure \ref{fig:teaser}.

\subsection{Screened Poisson equation with Hessian regularization}
Inspired by SPSR and SSD, we propose a new energy formulation for the surface reconstruction problem. The energy formulation is defined as:
\begin{equation}
  \resizebox{\linewidth}{!}{
    \begin{minipage}{\linewidth}
      \begin{align*}
        E(f) = E_N(f)+ \lambda_1 E_H(f) + \lambda_2 E_P(f),
        \tag{\theequation}
      \end{align*}
    \end{minipage}
  }
\end{equation}
\noindent where $E_N(f)$, $E_H(f)$, and $E_P(f)$ are the normal alignment, Hessian regularization, and point constraint energy terms, respectively, and $\lambda_1, \lambda_2 \in \mathbb{R}_{\ge 0}$ are weighting factors.

\begin{equation}
  \resizebox{\linewidth}{!}{
    \begin{minipage}{\linewidth}
      \begin{align*}
        E_N(f) = \iiint_{\Omega} \left\| \nabla f - \vec{\mathbf{N}} \right\|_2^2 \mathrm{d}V,
        \tag{\theequation}
      \end{align*}
    \end{minipage}
  }
\end{equation}
\noindent where $\nabla f$ is the gradient of the implicit field $f$, $\vec{\mathbf{N}}$ is the predicted unit normal field, and the integral is over the 3D domain $\Omega$.

In Eq (2), the first term $E_N(f)$ is the integration of the L2-norm of the difference between the gradient $\nabla f$ and the normal field $\vec{\mathbf{N}}$ in the 3D domain of $\Omega$, which measures the alignment between the surface and the level of the normal field. By promoting non-zero values exclusively at the shape boundary, we encourage $\vec{\mathbf{N}}$ to generate a reconstructed field that closely resembles a smoothed indicator function. When integrated across the entire domain, this term contributes to an overall estimation of the shape's orientation, with a strong regularization effect imposed by the normal field. This term is also known as the normal consistency term.
\begin{equation}
  \resizebox{\linewidth}{!}{
    \begin{minipage}{\linewidth}
      \begin{align*}
        E_H(f) = \iiint_{\Omega} \left\| H(f) \right\|_F^2 \mathrm{d}V,
        \tag{\theequation}
      \end{align*}
    \end{minipage}
  }
\end{equation}
\noindent where $H(f)$ is the Hessian matrix of $f$ and $\|\cdot\|_F$ denotes the Frobenius norm.

\revised{The second term, $E_H(f)$, is a Hessian energy term that regularizes the smoothness of the implicit field $f$. Minimizing the Frobenius norm of the Hessian matrix, $\left\| H(f) \right\|_F^2$, encourages the field's second derivatives to be close to zero. This property is characteristic of a true Signed Distance Function (SDF), which has a constant gradient magnitude of one, and therefore a zero Hessian. This regularization is particularly crucial for our Galerkin framework; it ensures that the implicit function remains smooth and avoids spurious oscillations in regions far from the input points $\mathcal{P}$, a common challenge when using global basis functions. It acts as a powerful smoothness prior on the entire domain $\Omega$, complementing the data-fitting terms without needing to estimate unstable geometric attributes like curvature directly from the sparse input points.}
\begin{equation}
  \resizebox{\linewidth}{!}{
    \begin{minipage}{\linewidth}
      \begin{align*}
        E_P(f) = &\sum_{p \in \mathcal{P}} \left( f(p) - \xi(p) \right)^2,
        \tag{\theequation}
      \end{align*}
    \end{minipage}
  }
\end{equation}

The third term $E_P(f)$ of Eq (1) is the point constraints term, which measures the discrepancy between the implicit function $f$ and the observed point cloud $\mathcal{P}$. The point constraints term is given by the L2-norm of the difference between the predicted values $\xi(p)$ and the observed values $f(p)$. By minimizing this term, we can obtain a surface that is well-aligned with the input data. This term ensures that the surface closely matches the input points, acting as a strong constraint. Particularly in regions with intricate geometric details, these point constraints play a crucial role in achieving a tight fit, thereby enhancing the clarity and accuracy of surface details. The use of $\xi(p)$ primarily serves the purpose of denoising by allowing points to be located outside the surface. So the whole formulation could be expressed as:
\begin{equation}
  \label{energy}
  \resizebox{\linewidth}{!}{
    \begin{minipage}{\linewidth}
      \begin{align*}
        \medmuskip=0.5mu 
        \thinmuskip=0.5mu 
        \thickmuskip=0.5mu 
        E(f) =  \iiint_{\Omega} \left\| \nabla f - \vec{\mathbf{N}} \right\|_2^2 \mathrm{d}V 
        + \lambda_1 \iiint_{\Omega} \left\| H(f) \right\|_F^2 \mathrm{d}V& \\
        + \lambda_2 \sum_{p \in \mathcal{P}} \left( f(p) - \xi(p) \right)^2&.
        \tag{\theequation}
      \end{align*}
      \label{eq:energy}
    \end{minipage}
  }
\end{equation}
This formulation allows for a balanced integration of learned priors and data fitting. The regularization term in the equation ensures that the square norm of the Hessian matrix is close to zero, resulting in a nearly constant gradient for the function away from the data points. In the vicinity of the data points, the contribution from the data energy terms dominates the total energy, while the regularization energy term plays a bigger role away from the data points, leading to a constant gradient vector field $\nabla f$. The Hessian regularization term encourages the surface to be smooth, while the signed distance error term encourages the surface to be as close as possible to the input point cloud. The entire formulation effectively achieves a balance between the model's data fitting capability and its generalization ability.

\subsection{Discretized implicit field on sparse voxel hierarchies}
\revised{Instead of converting point clouds into dense voxel grids or octree structures, we use a sparse voxel hierarchy required by our analytic solver. Local support of basis functions centered at voxel centroids yields a sparse, well‑conditioned linear system $\mathbf{A}\boldsymbol{\alpha}=\boldsymbol{b}$ solved directly; computation and memory scale with the number of active voxels via $\Omega=\bigcup Rgn(v)$; a sparse voxel backbone network selects and parameterizes active cells, predicting $\boldsymbol{m}_k^{\mathcal{S}}$ to inject learned priors while preserving the linear formulation. This design reduces cost during both learning and solving and enables coarse‑to‑fine control where fine levels add details only where data exist.}

Given a set of multi-layer sparse voxels $\mathcal{V} :=\{ \mathcal{V}^1, \mathcal{V}^2, , ...,\mathcal{V}^{\mathcal{S}} \}$ where $\mathcal{S}$ is the number of layers. Each voxel grid $\mathcal{V}^{\mathcal{S}}$ is a regular grid of voxels $\{v_k^\mathcal{S}|k=1,2,...\}$ with a fixed size $2^{\mathcal{S}-1}$ , where $b$ is the base voxel size. The voxel grid $\mathcal{V}^{\mathcal{S}}$ is strictly contained within $\mathcal{V}^{\mathcal{S}+1}$. Given a scaffold, we dynamically set the integration domain $\Omega$ as $\Omega:= \bigcup_\mathcal{S}\bigcup_{v_k^\mathcal{S}\in \mathcal{V}^\mathcal{S}} Rgn(v_k^\mathcal{S})$, where $Rgn(v)$ represents the enlarged region spanned by the voxel $v$ with a factor of $3^3$. 
To ensure an efficient solution to the problem described in equation (2) and maintain a high level of flexibility for surface representation, we define the target implicit function as follows:
\begin{equation}
  \begin{aligned}
  &f(\boldsymbol{p}) := \sum_{\mathcal{S}} \sum_{k} \alpha_{k}^{\mathcal{S}} \mathcal{B}_{k}^{\mathcal{S}}(\boldsymbol{p}), \\
  &\text{ with } \mathcal{B}_{k}^{\mathcal{S}}(\boldsymbol{p}) := B_{k}^{\mathcal{S}}\left(\frac{\boldsymbol{p}-Cntr(v_{k}^{\mathcal{S}})}{2^{\mathcal{S}-1}}\right).
  \end{aligned}
\end{equation}
The subscript $k$ represents the sum over all voxels in $\mathcal{V}^{\mathcal{S}}$, and $Cntr(v)$ denotes the centroid coordinate of voxel $v$. The function $B_{k}^{\mathcal{S}}: \mathbb{R}^3\rightarrow \mathbb{R}$ is the basis function that remains fixed during energy minimization. It exhibits spatial variation and, when multiplied by the coefficient $\alpha_{k}^{\mathcal{S}}\in\mathbb{R}$, describes the entire surface.

In contrast to the fixed Bézier tensor basis used in SPSR, we employ a parameterized family of basis functions that allows us to incorporate learned priors into the solution. The parameters of these basis functions are predicted by our upstream network, enabling the solver to flexibly incorporate inductive biases from the learned module.
\begin{equation}
  \resizebox{\linewidth}{!}{
    \begin{minipage}{\linewidth}
      \begin{align*}
        B_{k}^{s}(\boldsymbol{p})& :=B(\boldsymbol{p};\boldsymbol{m}_{k}^{s}):=\begin{cases}\prod_{a\in\{x,y,z\}}b^{\mathrm{a}}(a;\boldsymbol{m}_{k}^{s}),&\boldsymbol{p}\in\Omega_{B}\\0,&\boldsymbol{p}\notin\Omega_{B}\\\end{cases},  \\
        b^{\mathrm{x}}(x;\boldsymbol{m}_{k}^{s})& :=\sum_{u}m_{u}^{x}q_{u}^{x}(x),\quad m_{u}^{x}\in\mathbb{R},\quad q_{u}^{x}:\mathbb{R}\mapsto\mathbb{R}, 
        \tag{\theequation}
      \end{align*}
      \label{eq:basis}
    \end{minipage}
  }
\end{equation}
where $\boldsymbol{p}=[x, y, z]$ denotes three-dimensional space with axes $x$, $y$, and $z$, and the domain $\Omega_x := [-1.5, 1.5]$, which represents the support of the basis. We define functions $b^x(x)$, $b^y(y)$, and $b^z(z)$. And $q_u$ are elementary functions that are blended by weights $m_u$s, which vary across voxels. The vectorized form $\boldsymbol{m}_k^{\mathcal{S}} := [\ldots,m_u,\ldots]^T$ parameterizes the continuous function $B_{k}^{\mathcal{S}}$.
For more details about the basis function, please refer to the Appendix~\ref{subsec:basis_function}.

Utilizing the Euler-Lagrange equation (the detailed derivations of which are provided in the Appendix~\ref{subsec:optimal_surface_fitting}) and the aforementioned discretization scheme, we can determine the optimal surface fitting by solving the equation in closed-form. This involves finding the best possible linear coefficients $\boldsymbol{\alpha} := [..., \alpha_k^\mathcal{S}, ...]T$ for the implicit field $f$. 
\begin{equation}
  \resizebox{\linewidth}{!}{
    \begin{minipage}{\linewidth}
      \begin{align*}
        \mathbf{A}\boldsymbol{\alpha} & =b,  \\
        \mathrm{with}\quad\mathbf{A}_{(\bar{k},\bar{l})}& =\iiint_{\Omega}\nabla\mathcal{B}_{\bar{k}}^{\top}\nabla\mathcal{B}_{\bar{l}}\mathrm{d}V+\lambda_2\sum_{\bar{l}}\alpha_{\bar{l}}\iiint_{\Omega}\Delta\mathcal{B}_{\bar{k}}\cdot\Delta \mathcal{B}_{\bar{l}}\mathrm{d}V \\
        &+\lambda_1\sum_{\bar{l}}\alpha_{\bar{l}}\sum_{p\in\mathcal{P}}\mathcal{B}_{\bar{k}}(\boldsymbol{p})\mathcal{B}_{\bar{l}}(\boldsymbol{p}), \\
        b_{(\bar{k})}&= \iiint_{\Omega}\nabla\mathcal{B}_{\bar{k}}^{\top}\vec{\mathbf{N}}\mathrm{d}V+\lambda_1\sum_{p\in\mathcal{P}}\mathcal{B}_{\bar{k}}(\boldsymbol{p})\xi(\boldsymbol{p}),
        \tag{\theequation}
      \end{align*}
      \label{eq:closed-form}
    \end{minipage}
  }
\end{equation}

In this equation, the subscripts $\overline{k} $ and $\overline{l}$ represent the row and column indices of the matrix, respectively. The optimization problem takes into account all voxels across all scales, with $\overline{k} $ and $\overline{l}$ indexing the product of the subscripts $k$ and $s$ ($i.e.,\mathcal{B}_{k}^{\mathcal{S}}$ is represented as $\mathcal{B}_{\overline{k}}$ as previously defined).

\subsection{Multi-layer convolution}
With an adaptive sparse neural network applied to sparse voxel hierarchies, the goal is to predict key attributes required by the surface fitting solver. This section primarily focuses on elucidating how the multi-layer convolution is employed to learn valuable priors from various layers. 
\begin{figure}[!t]
  \centering
  \subfloat[]{\includegraphics[width=1in]{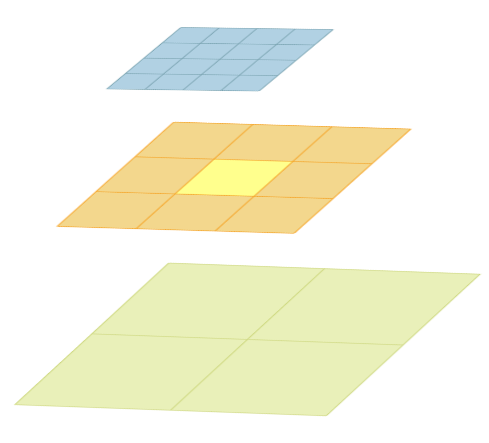}%
  \label{origin kernel}}
  \hfil
  \subfloat[]{\includegraphics[width=1in]{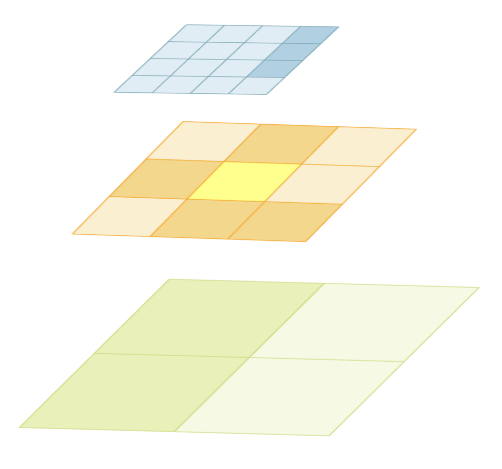}%
  \label{filtered kernel}}
  \hfil
  \subfloat[]{\includegraphics[width=1in]{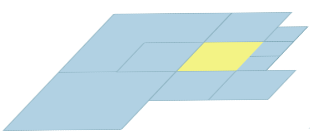}%
  \label{multi-layer kernel}}
  \caption{When focusing on the two-dimensional special case for ease of understanding, the original multi-layer convolutional kernel consists of 29 elements, derived from layers of different sizes from three neighboring hierarchies, as illustrated in \ref{origin kernel}. During the convolution operation, the voxels within the convolution kernel are filtered based on the number of points computed inside the voxel, as shown in \ref{filtered kernel}. This results in a convolutional kernel of the form \ref{multi-layer kernel}. The actual convolution process, however, occurs in three dimensions, where the original multi-layer convolutional kernel consists of 107 voxels, originating from three adjacent hierarchies.}
  \label{fig_sim}
\end{figure}

To handle the adaptive grids with irregular spacing in the input point cloud, traditional regular 3D convolutions and even sparse convolutions are not suitable. Regular 3D convolutions can be expressed in the following formulation:
\begin{equation}
  \resizebox{\linewidth}{!}{
    \begin{minipage}{\linewidth}
      \begin{align*}
        F_i = \sum_{j\in\text{NB}(i)}\boldsymbol{W}(\Delta p_{ij})\times F_j,
        \tag{\theequation}
      \end{align*}
    \end{minipage}
  }
\end{equation}
Where $F_i$ and $F_j$ are the input and output features of the network, $p_{ij}$ are the relative position of voxels $v_i$ and $v_j$, $\boldsymbol{W}$ is the weight matrix of the network, $\text{NB}(x)$ is the set of neighboring voxels of $x$. The regular 3D convolution expects data to be represented at regularly spaced points, while sparse convolutions assume regular grids and do not account for the irregular spacing found in adaptive grids. Both of them need to be addressed on the same layer of sparse voxels, which means that the input features are detailed when the input point clouds are dense around the output voxel, while input features are rough when the input point clouds are sparse. In order to address the non-uniform distribution of point clouds and ensure a balanced aggregation of point information during each convolution, we introduce a multi-layer convolution approach at the voxel level. Figure \ref{fig_sim}(a) shows an example of the kernels. This allows for 3D convolutions to be performed on voxels across different layers, and can be expressed as:

\begin{equation}
  \resizebox{\linewidth}{!}{
    \begin{minipage}{\linewidth}
      \begin{align*}
        F_i = F_i + \sum_{j\in\text{NB}(i)}\boldsymbol{W}(\Delta p_{ij})\times \mathcal{C} (F_j^{down},F_j^{cur}, F_j^{up}),
        \tag{\theequation}
      \end{align*}
    \end{minipage}
  }
\end{equation},
where $\mathcal{C}$ is a convolutional kernel function that takes into account the spatial relationships between different layers. 
\begin{equation}
  \resizebox{\linewidth}{!}{
    \begin{minipage}{\linewidth}
      \begin{align*}
        \medmuskip=0.5mu 
        \thinmuskip=0.5mu 
        \thickmuskip=0.5mu 
        \mathcal{C} (F_j^{down},F_j^{cur}, F_j^{up})& :=
        \begin{cases}F_j^{down},&\boldsymbol{N}_{v_j}>\boldsymbol{N}_{max}\\
          F_j^{cur},&\boldsymbol{N}_{min}<\boldsymbol{N}_{v_j}<\boldsymbol{N}_{max}\\
          F_j^{up},&\boldsymbol{N}_{v_j}<\boldsymbol{N}_{min}\\\end{cases},
      \end{align*}
    \end{minipage}
  }
\end{equation}
Where $\boldsymbol{N}_{v_j}$ is the number of neighboring voxels of $v_j$ in the down, current, and up directions, respectively. The convolutional kernel has 27 elements, one for each combination of neighboring voxels. During the convolution operation, it is possible to filter the voxels within the convolution kernel based on a pre-calculated number of points. Voxel features that meet the criteria will be retained, and ultimately, the features from 67 voxels in the upper and lower layers will be aggregated into a convolution kernel of size 27, provided that both upper and downer layers exist. We have incorporated residual connections to mitigate the issue of overfitting during the learning process of voxels across different layers.

\subsection{Point-Voxel Attention}

\revised{Voxel-based convolutional methods typically aggregate point features within each voxel into a single feature vector. This aggregation, while efficient, causes a loss of fine-grained geometric detail by discarding the spatial distribution of points within a voxel. Such information loss is detrimental for high-fidelity surface reconstruction. To address this, we introduce a Point-Voxel Attention mechanism as a complementary module to our sparse convolutional backbone. This mechanism refines voxel features by attending to the individual points they contain, thereby re-introducing crucial intra-voxel positional information.}

\revised{Let $F_v \in \mathbb{R}^{C_v}$ be the feature of a voxel $v$, and let $\{\boldsymbol{p}_j\}_{j=1}^{N_v}$ be the set of $N_v$ points contained within this voxel, each with its feature $F_{p_j} \in \mathbb{R}^{C_p}$. Our goal is to update the voxel feature $F_v$ by aggregating information from its constituent points in a spatially-aware manner.}

\revised{The core idea is to compute attention weights for each point based on its features and relative position within the voxel. The updated voxel feature is then a weighted sum of point-derived values. The process is as follows:}

\revised{First, we project the point and voxel features into a common embedding space to derive query ($\boldsymbol{q}$), key ($\boldsymbol{k}$), and value ($\boldsymbol{u}$) vectors. The query and value vectors are derived from point features, while the key vector is derived from the voxel feature:
\begin{equation}
    \boldsymbol{q}_j = \phi_q(F_{p_j}), \quad \boldsymbol{u}_j = \phi_u(F_{p_j}), \quad \boldsymbol{k}_v = \psi_k(F_v)
\end{equation}
where $\phi_q$, $\phi_u$, and $\psi_k$ are implemented as simple multi-layer perceptrons (MLPs).}

\revised{To incorporate crucial intra-voxel positional information, we compute a positional encoding $F_{\text{pos}_j}$ from the relative coordinates $\boldsymbol{\delta}_j = \boldsymbol{p}_j - Cntr(v)$ of each point $\boldsymbol{p}_j$ with respect to the voxel center $Cntr(v)$:
\begin{equation}
    F_{\text{pos}_j} = \phi_{\text{pos}}(\boldsymbol{\delta}_j)
\end{equation}
where $\phi_{\text{pos}}$ is another MLP.}

\revised{The attention weight $\alpha_j$ for each point $\boldsymbol{p}_j$ is then computed based on the compatibility between the point's query and the voxel's key, modulated by the positional encoding:
\begin{equation}
    w_j = \phi_w(\boldsymbol{k}_v - \boldsymbol{q}_j + F_{\text{pos}_j})
\end{equation}
Here, $\phi_w$ is an MLP that maps the compatibility score to a single scalar. The weights are normalized across all points within the same voxel using the softmax function. Finally, the new voxel feature $F'_v$ is computed by aggregating the point-based values, also enhanced with positional encoding, using the calculated attention weights:
\begin{equation}
    F'_v = \sum_{j=1}^{N_v} \frac{\exp(w_j)}{\sum_{i=1}^{N_v} \exp(w_i)} (\boldsymbol{u}_j + F_{\text{pos}_j})
\end{equation}
This updated feature $F'_v$ is then added to the original voxel feature $F_v$ via a residual connection, enriching it with fine-grained, position-aware information from the raw point cloud. This mechanism allows the network to learn a more precise representation of the underlying surface geometry, complementing the broader contextual understanding provided by the sparse convolutions.}

\section{Experiment}
\label{sec:exp}
In this section, we introduce the experimental settings, including the parameters and evaluation metrics used in our experiments. We then present the results of our method at different scales, including object-level datasets and indoor scene datasets. We also conduct ablation studies and analyze the time and memory efficiency of our method.
\subsection{Implementation Details}
We implement our method based on a high-performance point cloud inference engine, Torchsparse \cite{tang2022torchsparse} to accelerate the computation of neighborhood queries and sparse matrix operations, the sparse multi-layer convolutional network shares its neighborhood map with the linear solver. We use an AWSGrad optimizer \cite{reddi2018convergence} with a learning rate of 0.001 and a batch size of 1. In our experiments, we set the parameters as follows: $\lambda_1= 64.0$, $\lambda_2= 3.0$, with $S$ representing 4 layers of voxel hierarchies and $M$ varying across different datasets. 
\revised{The total training loss, $\mathcal{L}_{\text{total}}$, is a weighted sum of losses for structural prediction, voxel-level normal supervision, and surface geometry. The structure loss, $\mathcal{L}_{\text{struct}}$, is a multi-level cross-entropy loss that supervises voxel occupancy at different depths $d$ of the hash tree. The voxel normal loss, $\mathcal{L}_{\text{vox\_normal}}$, uses an L1 distance to enforce consistency between predicted and ground-truth normals on the grid. The surface geometry loss combines a surface loss $\mathcal{L}_{f}$, which forces the implicit function $f$ to be zero on the surface, and a gradient loss $\mathcal{L}_{\text{grad}}$, which aligns the predicted gradient $\nabla f$ with the ground-truth surface normal $\mathbf{n}_{\text{gt}}$ using a dot product.
The complete loss function can be summarized by the following equation:}

\begin{align}
\mathcal{L}_{\text{total}} &= \sum_d w_{\text{struct}} \cdot \mathcal{L}_{\text{struct},d} + \sum_d w_{\text{vox}} \cdot \mathcal{L}_{\text{vox\_normal},d} \nonumber \\
&\quad + \mathbb{I}_{\text{full}} \cdot w_{\text{surf}} \left( \mathcal{L}_{f} + w_{\text{grad}} \mathcal{L}_{\text{grad}} \right),
\end{align}
Where $d$ is the depth on voxel hierarchy $S$. $\mathbb{I}_{\text{full}}$ is an indicator function that is 0 during the fist training phase and 1 otherwise.
$w_{\text{struct}}$, $w_{\text{vox}}$, $w_{\text{surf}}$, and $w_{\text{grad}}$ are the respective weights for each loss component from the hyperparameters.
The efficacy of these parameter selections will be explored further in our forthcoming ablation studies. Supervision is provided using 10000 points on Matterport3D and 1000/3000 points on ShapeNet dataset. We train our model on an NVIDIA V100 GPU with 32GB memory and an Intel Xeon E5-2698 v4 CPU.

\subsection{Evaluation Metrics}
We evaluate the reconstruction quality using several standard metrics:
F-Score (\%)~($\uparrow$), which balances precision and recall;
Chamfer Distance~($\downarrow$), the L1 distance between prediction and ground truth;
Normal Consistency (\%)~($\uparrow$), the alignment of surface normals;
and Intersection-over-Union (IoU)~($\uparrow$), the volumetric accuracy for watertight meshes.

\subsection{Object-level Reconstruction}
In the context of object-level evaluations, we utilize the ShapeNet datasets \cite{changshapenet}, which comprises a comprehensive collection of over 50,000 unique three-dimensional models specifically curated for deep learning applications. Consistent with established protocols in the field, we employ the train-validation-test partition proposed by \cite{choy20163d}, which encompasses thirteen distinct object categories. A total of 8,751 previously unseen objects across these categories are employed for benchmarking purposes. Following the methodology of \cite{mescheder2019occupancy}, we generate virtually-fused watertight meshes to serve as ground truth, as opposed to utilizing the original raw meshes that may not adhere to manifold constraints. All meshes are normalized to fit within a unit cube. Our experimental framework is structured around three distinct scenarios: 'No noise, 1K points', 'Small noise, 3K points', and 'Large noise, 3K points'. Here, 'small noise' and 'large noise' pertain to Gaussian noise perturbations superimposed on the input point sets, with standard deviations of 0.005 and 0.025, respectively. 

Given the approximately uniform geometric variations observed in the ShapeNet objects, with few instances of less detailed surfaces such as planes, we deem the implementation of an adaptive voxel grid structure unnecessary for this dataset. Consequently, we set the maximum adaptive depth, $M$, to 1, signifying that voxels at scales $s > 1$ are not preserved in their original form. In constructing the ground-truth voxel grid for structural supervision $(c_k^s)_gt$, we ensure that each point on the ground-truth surface is encapsulated by a voxel at all scales. A voxel size of $b = 0.02$ has been determined sufficient for accurately modeling the shapes in question.

In our study, we compare our approach against several established baselines including the Screened Poisson Surface Reconstruction (SPSR) as detailed in \cite{kazhdan2013screened}, Convolutional Occupancy Networks (ConvONet) reported by \cite{peng2020convolutional}, Neural Galerkin(NGSolver) by \cite{huang2022neural}, and NKSR from \cite{huang2023neural}. All learnable models were retrained from the ground up.

\begin{figure*}[ht]
    \centering
    \includegraphics[width=\textwidth]{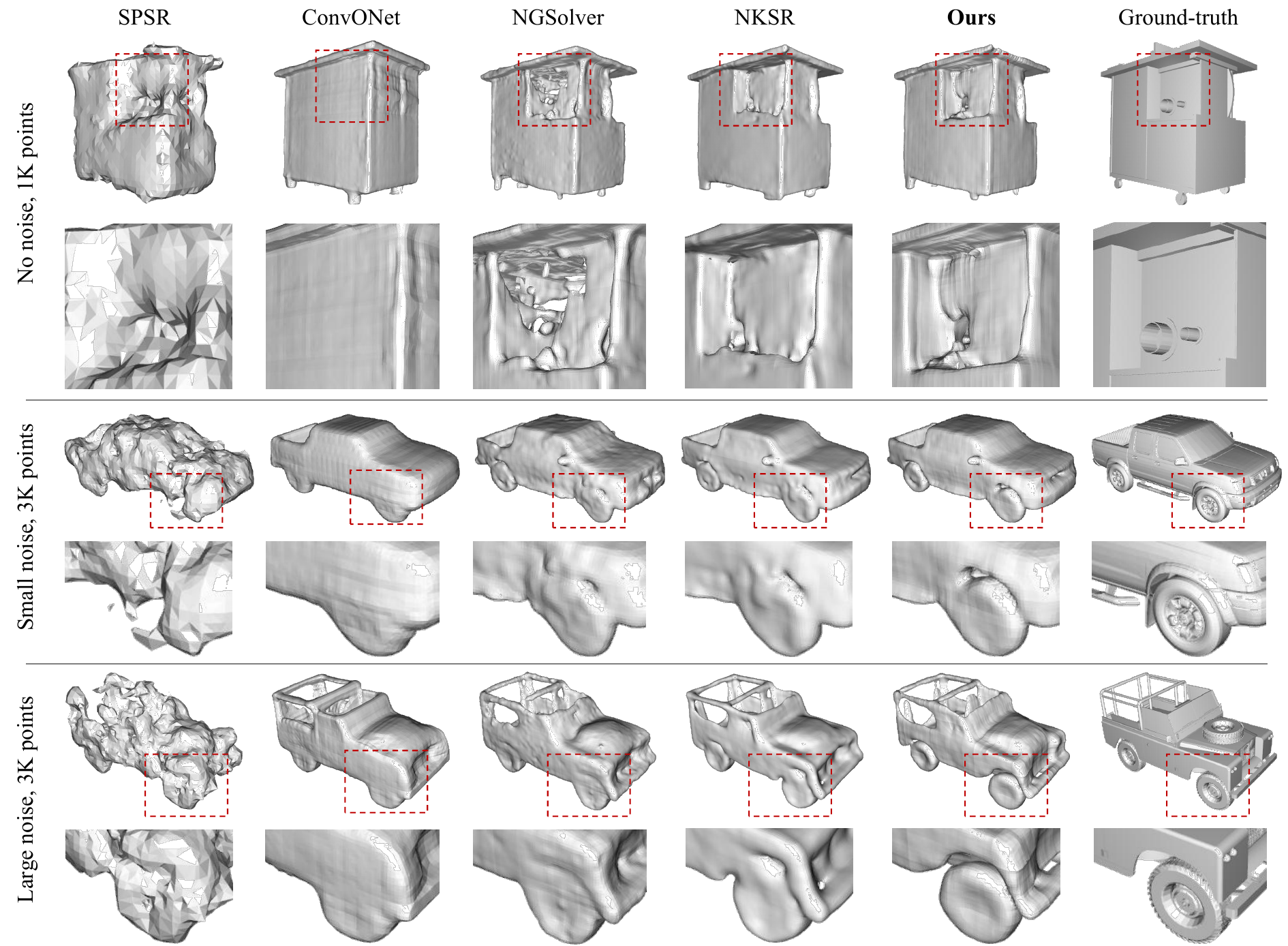}
    \caption{Here are the results on the ShapeNet dataset\cite{changshapenet} across three different settings. Methods SPSR utilize points normals, while others do not. Notably, our method demonstrates a capability to accurately reconstruct fine geometric details (such as the car wheels).}
    \label{fig:ShapeNet}
\end{figure*}

\begin{table*}[ht]
    \centering
    \caption{ShapeNet Object-level Reconstruction Results}
    \label{ShapeNet}
    \resizebox{\textwidth}{!}{
    \begin{tabular}{|c|c|c|c|c|c|c|c|c|c|c|c|c|}
    \hline
    Method & \multicolumn{4}{c|}{No noise, 1K points} & \multicolumn{4}{c|}{Small noise, 3K points} & \multicolumn{4}{c|}{Large noise, 3K points} \\ \cline{2-13} 
    & Chamfer ↓ & F-score ↑ & Normal C. ↑ & IoU ↑ & Chamfer ↓ & F-score ↑ & Normal C. ↑ & IoU ↑ & Chamfer ↓ & F-score ↑ & Normal C. ↑ & IoU ↑ \\ 
    \hline
    ConvONet & 6.38 & 88.2 & 92.0 & 81.6 & 4.65 & 93.4 & 93.5 & 87.4 & 7.69 & 80.9 & 90.7 & 77.8\\
    NGSolver & 3.07 & 97.1 & 94.4 & 91.3 & 3.03 & 98.4 & 94.7 & 91.3 & 5.24 & 89.4 & 91.6 & 82.1 \\
    NKSR & \textbf{2.54} & 97.6 & 94.3 & 92.7  & \textbf{2.57} & 98.7 & 94.6 & 91.8  & \textbf{5.01} & 89.8 & 91.4 & 82.1  \\
    Ours & 2.71 & \textbf{98.5} & \textbf{95.9} & \textbf{94.0} & 2.77 & \textbf{99.1} & \textbf{96.0} & \textbf{93.0} & 5.03 & \textbf{90.4} & \textbf{92.1} & \textbf{82.7}  \\
    \hline
    \end{tabular}
    }
\end{table*}
The results of these comparisons are summarized in Table \ref{ShapeNet} and visually presented in Figure \ref{fig:ShapeNet}. The non-learnable modules SPSR struggle to reconstruct detailed geometries from sparse datasets as small as 1K points and are notably affected by noise, which leads to distorted, bulging geometrical outputs. Additionally, our method demonstrates superior reconstruction accuracy over ConvONet. This is notable since ConvONet infers implicit fields directly without any guarantees of fit.

Our method's voxel grid resolution is comparable to that of ConvONet, which uses neural-network-parameterized local implicit fields. However, our approach achieves approximately a 9$\%$ higher F-score, underscoring the robust representational capabilities of our elementary-function-based bases. Unlike NGSolver, which tackles a screened variant of the Poisson equation using a discrete grid, our method employs a Hessian-based regularizer to enforce smoothness, thereby capturing more detailed and faithful models of the shapes with a highly accurate SDF gradient field. 

\subsection{Room-level Reconstruction}
The scalability of our approach is further validated on the Matterport3D dataset ~\cite{chang2017matterport3d}, which comprises real-world scans of indoor environments captured using a panoramic depth camera. This dataset includes a total of 90 buildings. Adhering to the prescribed train/test division, we further segment each building into discrete rooms, each of which is reconstructed and assessed independently. The test dataset encompasses 399 rooms. For supervisory data generation, we employ the methodology outlined in § 5.3, which generates larger-scale voxels in planar areas such as walls and floors. We configure our system with $M$ = 2 and $b$ = 0.012.

\begin{figure*}[t]
    \centering
    \includegraphics[width=\textwidth]{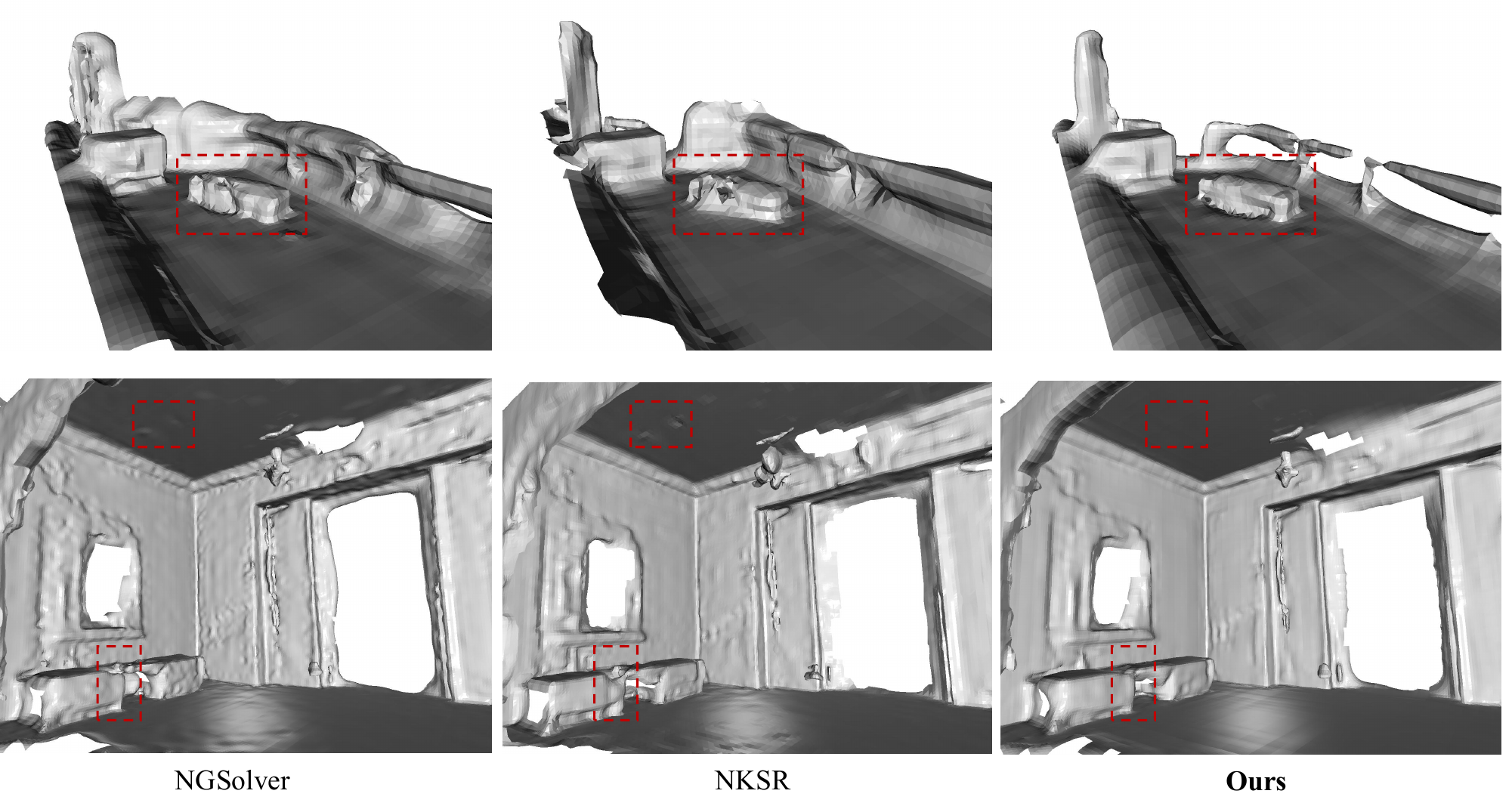}
    \caption{Qualitative results on the Matterport3D dataset. Our method demonstrates superior performance in reconstructing large-scale indoor scenes, excelling at capturing fine details, such as the chair highlighted in the red box, while simultaneously ensuring overall structural integrity by eliminating noise from surfaces like walls and floors.}
    \label{fig:matterport_comparison}
\end{figure*}

While the indoor scans consist solely of single-sided surfaces, our solver's tendency to produce watertight geometries results in the presence of minimal floaters at the domain boundaries. Utilizing a truncated domain $\Omega$, these floaters are contained within a small area and are strategically prevented from emerging in areas distant from the surface. In our evaluations, we implement a k-nearest neighbors (k-NN) strategy to eliminate vertices on the produced mesh that are excessively distant from the input points.
\begin{table}[ht]
    \centering
    \caption{Matterport3D Room-level Reconstruction Results}
    \label{room}
    \begin{tabular}{lccc}
    \toprule
    Method & \multicolumn{3}{c}{No noise, 10K points} \\ \cmidrule(lr){2-4}
    & Chamfer $\downarrow$ & F-score $\uparrow$ & Normal C. $\uparrow$ \\ 
    \midrule
    ConvONet & 6.21 & 88.9 & 92.3 \\
    NGSolver & 3.79 & 93.5 & 94.3  \\
    NKSR & \textbf{2.60} & 96.4 & \textbf{96.5}  \\
    Ours & 3.07 & \textbf{97.9} & 94.9 \\
    \bottomrule
    \end{tabular}
\end{table}
For quantitative evaluations, we adhere to established protocols and reduce the input point cloud to 10K points. We benchmarked our method against ConvONet, NGSolver and NKSR. The outcomes of the comparative analysis are meticulously presented in Table \ref{room} and visually represented in Figure \ref{fig:matterport_comparison}. Our approach consistently surpasses all comparative benchmarks in F-score, the primary metric for reconstruction accuracy, delivering precise and detailed reconstructions at the room level with intricate geometric nuances. Notably, our methodology demonstrates a natural benefit wherein the accuracy of reconstruction progressively improves as more input data points are provided, without requiring modifications to the network architecture—a characteristic not regularly observed in several contemporary learning-based models, such as ConvONet. 

\revised{\subsection{Evaluation on Surface Reconstruction Benchmark}
We further evaluate our method on the Surface Reconstruction Benchmark (SRB), comparing it against leading baselines. Following the standard protocol, all methods process an input of 3,000 points, and learning-based models are trained on ShapeNet. Table \ref{tab:srbbenchmark} summarizes the quantitative results. Our approach demonstrates superior performance, which underscores its strong generalization capabilities.}

\begin{figure*}[h!]
\centering
\includegraphics[width=\linewidth]{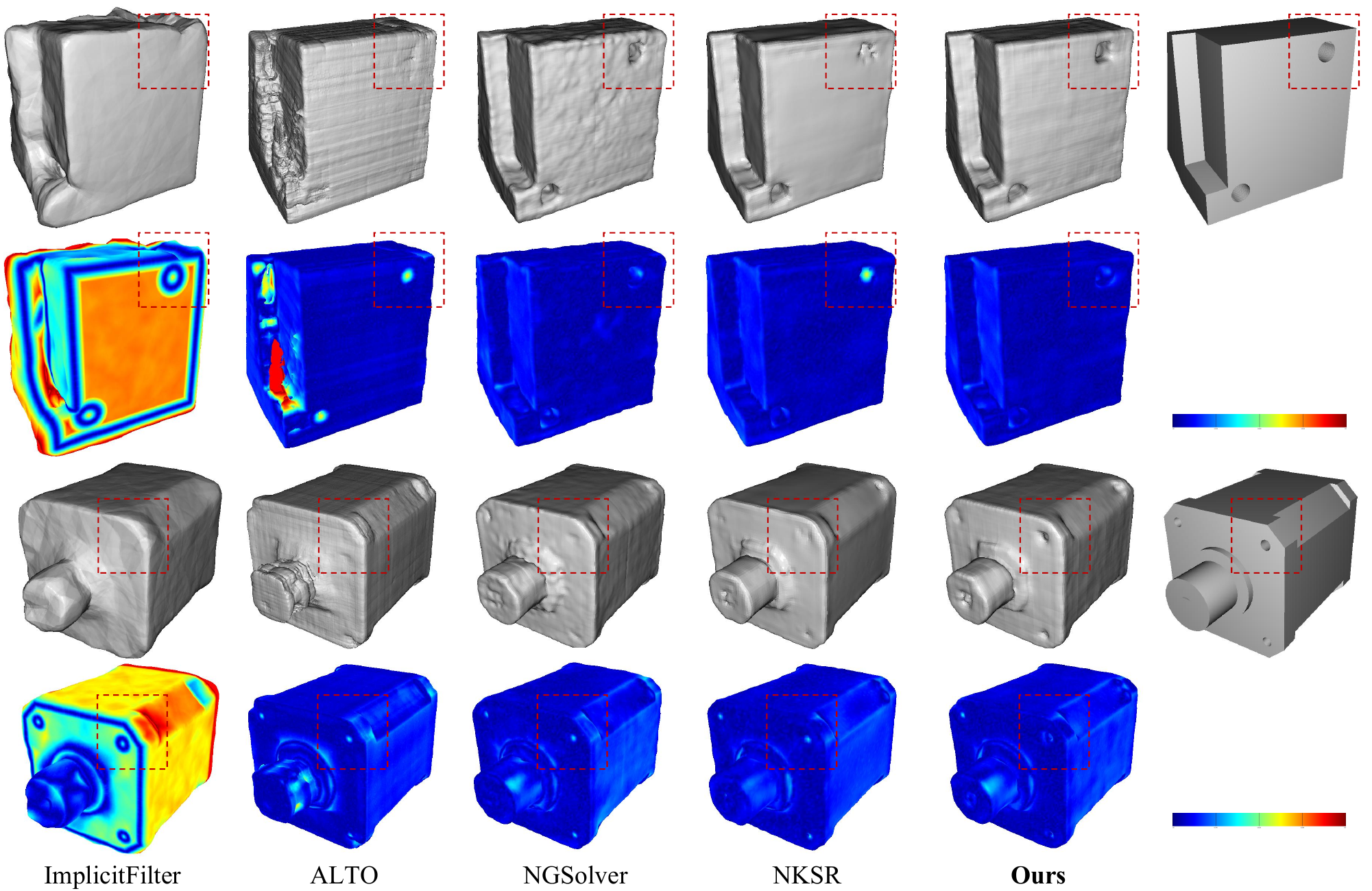}
\caption{Qualitative comparison on the Surface Reconstruction Benchmark (SRB). Our method achieves a better balance between surface smoothness and detail preservation. The error is visualized with a color map where blue indicates low error and red indicates high error.}
\label{fig:srb_comparison}
\end{figure*}

\revised{For qualitative assessment, as illustrated in Figure \ref{fig:srb_comparison}, we compare our method with state-of-the-art approaches like NKSR and NGSolver. NKSR tends to produce overly smooth surfaces, sacrificing fine geometric details. In contrast, NGSolver preserves details to an extent that results in a rough and noisy surface. Our method strikes a better balance, effectively capturing the overall object semantics while retaining crucial details.}

\begin{table}[h!]
\centering
\caption{Quantitative comparison on the Surface Reconstruction Benchmark (SRB). Best results are in \textbf{bold}.}
\label{tab:srbbenchmark}
\begin{tabular}{lccc}
\toprule
Method & Chamfer $\times 10^{-2}$ $\downarrow$ & F-Score (\%) $\uparrow$ & Normal C. $\uparrow$ \\
\midrule
OG-INR\cite{koneputugodage2023octree} & 8.988 & 11.1 & 59.1 \\
ImplicitFilter\cite{li2024implicit} & 3.764 & 31.9 & 79.6 \\
POCO\cite{boulch2022poco} & 1.255 & 73.4 & 92.0\\
ALTO\cite{wang2023alto} & 1.245 & 71.8 & 90.3 \\
NKSR\cite{huang2023neural} & 1.118 & 76.1 & 93.2 \\
NGSolver\cite{huang2022neural} & 1.105 & 76.9 & 93.0 \\
\midrule
\textbf{Ours} & \textbf{1.099} & \textbf{77.2} & \textbf{93.4} \\
\bottomrule
\end{tabular}
\end{table}

\subsection{Ablation Studies}
Ablation experiments are executed utilizing the ShapeNet and Matterport datasets. Specifically, for ShapeNet, the 'Small noise, 3K points' configuration is selected, focusing on its chair category for all phases: training, validation, and testing. Under identical parameter settings, our model, trained solely on this subset, demonstrates comparable performance to that trained on the comprehensive dataset (F-score 98.4 versus 98.6). Beyond the F-score and normal consistency metrics, we transition the chamfer distance metric from the L1-norm to the L2-norm (hereinafter referred to as 'L2-Chamfer') to enhance the visibility of differences, given its sensitivity to outliers. The L2 distance is adjusted by a factor of $10^5$ to facilitate clearer visualization.

\begin{table}[ht]
    \centering
    \caption{Performance Metrics for Various $\lambda_2$ Settings.}
    \label{lambda2}
    \begin{tabular}{cccc}
    \toprule
    $\lambda_2$ & Chamfer$\times 10^{-3}$ $\downarrow$ & F-score $\uparrow$ & Normal C. $\uparrow$ \\ \midrule
    0.0 & 3.00 & 98.2 & 93.7 \\ 
    2.0 & 2.92 & 98.4 & 93.8 \\ 
    3.0 & \textbf{2.55} & \textbf{98.5} & \textbf{93.9} \\ 
    6.0 & 2.91 & 98.4 & 98.9 \\ \bottomrule
    \end{tabular}
    \end{table}
\subsubsection{Energy formulation with Hessian regularization}
The balancing factor $\lambda_2 $, as defined in Equation \ref{energy}, is employed to ensure the smoothness of the SDF gradient field by constraining the norm of the Hessian matrix. In scenarios where $\lambda_2 $ equals zero, the approach reverts to a basic Screened Poisson solution, as described by \cite{kazhdan2013screened}, with a sole focus on the normal data, which in our implementation is predicted by the network. As depicted in TABLE \ref*{lambda2}, all metrics show improvement with $\lambda = 3 $ . These findings demonstrate the significant role of the Hessian regularization term in producing higher-quality geometrical constructs. It is essential to recognize that the elements in Equation \ref{energy}, aside from $\lambda_2$, are pivotal, offering critical guidance for determining the surface location and orientation. Notably, the absence of these terms in Equation (5), specifically as $\lambda$ approaches infinity, results in the non-convergence of our solver.

\subsubsection{Multi-Layer Convolution}
We conducted an ablation study to ascertain the efficacy of multi-layer convolution in the context of adaptive sparse neural networks applied to sparse voxel hierarchies for surface reconstruction tasks. The study comprised three distinct experimental setups: (1) a baseline model without multi-layer convolution, (2) a model integrating multi-layer convolution within the encoder, and (3) a model employing multi-layer convolution across both the encoder and decoder. The inclusion of multi-layer convolutions is intended to facilitate a more nuanced aggregation of point information across varying densities of input point clouds, thereby enhancing the model's ability to capture and utilize complex spatial hierarchies.

The performance of each configuration was rigorously evaluated against three critical metrics: Chamfer Distance, F-Score, and Normal Consistency, as shown in the table below. The results distinctly illustrate the substantial gains in performance metrics when multi-layer convolutions are implemented, particularly when incorporated throughout both the encoder and decoder. This configuration yielded the highest scores across all evaluated metrics, substantiating the proposed hypothesis regarding the advantages of multi-layer convolution in handling complex and non-uniform datasets.

\begin{table}[h]
    \centering
    \caption{Performance comparison of different configurations. \\
    MLC: Multi-Layer Convolution, Enc.: Encoder, Dec.: Decoder}
    \label{tab:performance}
    \begin{tabular}{cccc}
    \toprule
    Config. & Chamfer $\times 10^{-3}$ $\downarrow$ & F-score $\uparrow$ & Normal C. $\uparrow$\\ \midrule
    No MLC & 3.03 & 98.12 & 93.4 \\ 
    Enc. + MLC & 2.95 & 98.29 & 93.7 \\ 
    Enc. \& Dec. + MLC & \textbf{2.92} & \textbf{98.34} & \textbf{93.8} \\ \bottomrule
    \end{tabular}
\end{table}

\begin{table}[h!]
\centering
\caption{Ablation study of the Point-Voxel Attention (PVA) mechanism.}
\label{tab:ablation_pva}
\begin{tabular}{l|ccc}
\hline
Method & Chamfer $\times 10^{-3}$ $\downarrow$ & F-score $\uparrow$ & Normal C. $\uparrow$ \\
\hline
Ours (w/o PVA) & 3.17 & 98.2 & 94.5 \\
Ours (w/ PVA)  & \textbf{3.01} & \textbf{98.6} & \textbf{94.9} \\
\hline
\end{tabular}
\end{table}

\subsubsection{Point-Voxel Attention}

\revised{To validate the effectiveness of the Point-Voxel Attention mechanism, we conducted an ablation study by comparing the model's performance with and without this component. The results, summarized in Table~\ref{tab:ablation_pva}, demonstrate a clear improvement across all key metrics when the Point-Voxel Attention is integrated.}

\revised{Specifically, the inclusion of Point-Voxel Attention leads to a lower Chamfer Distance, a higher F-score, and improved Normal Consistency . This indicates that our attention mechanism effectively enhances the model's ability to capture fine-grained geometric details and produce more accurate surface reconstructions. The performance gains underscore the importance of aggregating features from both point and voxel representations for high-fidelity shape completion.}

\subsection{Computational Efficiency Analysis}

\revised{To evaluate the practical applicability of our method, we conducted a comprehensive analysis of computational time and memory usage across different configurations. All experiments were performed on a workstation equipped with an NVIDIA RTX 4090 GPU and 64GB RAM.}

\begin{table}[h]
    \centering
    \caption{Computational efficiency comparison across different methods.}
    \label{tab:efficiency}
    \begin{tabular}{ccc}
    \toprule
    Method  & Inference Time (s) & Memory Usage (GB) \\
    \midrule
    ImplicitFilter\cite{li2024implicit} & 30.3 & 2.1\\
    OG-INR\cite{koneputugodage2023octree} & 18.5 & 5.54 \\
    POCO\cite{boulch2022poco} & 5.3 & 4.6 \\
    ALTO\cite{wang2023alto}  & 4.5 & 3.8 \\
    Ours & 2.5 & 4.1 \\
    \bottomrule
    \end{tabular}
\end{table}

The efficiency analysis reveals that our method achieves competitive computational performance while maintaining superior reconstruction quality. Although the full configuration requires slightly more computational resources than the baseline, the increase is marginal compared to the significant quality improvements observed. Notably, our method demonstrates faster inference times compared to traditional methods like ImplicitFilter and OG-INR, while achieving substantially better reconstruction quality with a comparable memory footprint.

\section{Conclusion}
In this research paper, we introduce NeuralSSD, an innovative framework for 3D surface reconstruction that generates an implicit 3D surface from commonly available point cloud data. Through comprehensive experiments, we demonstrate that our method achieves a desirable balance between point fitting fidelity and strong generalization capabilities to different objects. We believe that our approach enables higher-definition outputs, demonstrates good generalization to unseen data, and makes deep-learning-based surface reconstruction more accessible for various applications. As part of future research, we aim to enhance the reconstruction quality by exploring alternative families of basis functions. Additionally, we plan to reduce the memory footprint to enable larger-scale reconstructions.


 
%

\cleardoublepage

\section{Biography Section}
\begin{IEEEbiography}
[{\includegraphics[width=1in,height=1.25in,clip,keepaspectratio]{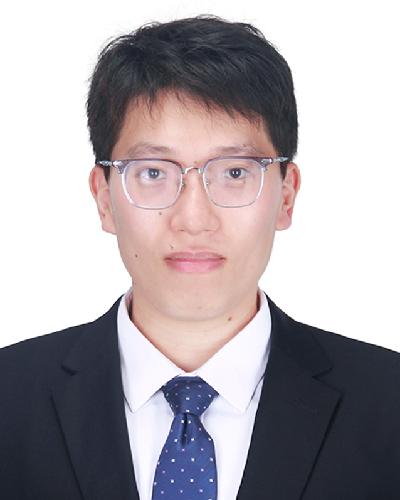}}]{Zi-Chen Xi}
is a Master's candidate at the Departemnt of Computer Science and Technology, Tsinghua University. He previously received his Bachelor's degree in Architecture from the School of Architecture at Tsinghua University in 2022. His research interests lie in 3D computer vision and computer graphics.
\end{IEEEbiography}

\begin{IEEEbiography}
[{\includegraphics[width=1in,height=1.25in,clip,keepaspectratio]{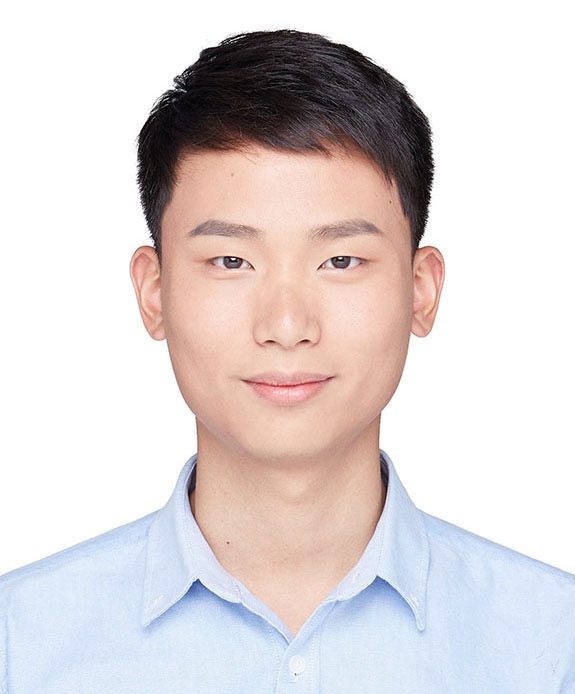}}]{Jiahui Huang}
is currently a research scientist at NVIDIA Toronto AI Lab led by Sanja Fidler. He received my Ph.D. in 2023 from the Graphics and Geometric Computing Group at Tsinghua University, China, advised by Shi-Min Hu. He was also a visiting researcher in the Geometric Computing group at Stanford University, led by Leonidas Guibas. His primary research interest lies in the joint field of 3D computer vision and graphics, including neural reconstruction, dynamic scene perception, and SLAM.
\end{IEEEbiography}

\begin{IEEEbiography}
[{\includegraphics[width=1in,height=1.25in,clip,keepaspectratio]{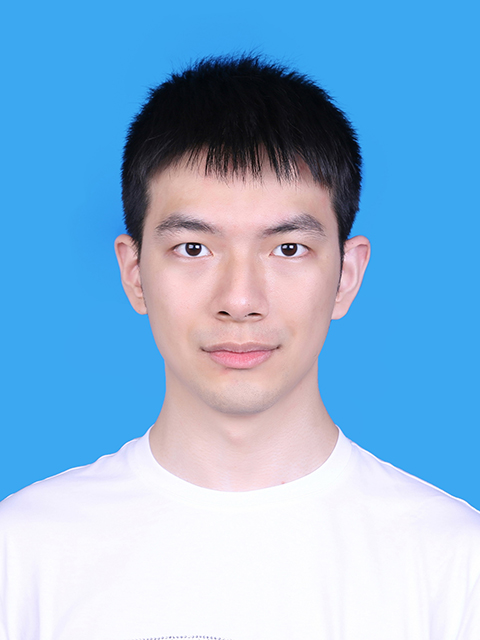}}]{Hao-Xiang Chen}
received his bachelor's degree in computer science from JiLin University in 2020. He is currently a PhD candidate in the Department of Computer Science and Technology, Tsinghua University. His research interests include 3D reconstruction and 3D computer vision.
\end{IEEEbiography}

\begin{IEEEbiography}
[{\includegraphics[width=1in,height=1.25in,clip,keepaspectratio]{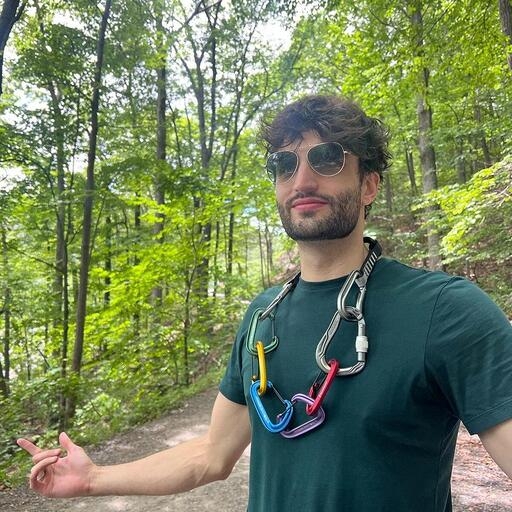}}]{Francis Williams}
is a senior research scientist at NVIDIA in NYC working at the intersection of computer vision, machine learning, and computer graphics. His research is a mix of theory and application, aiming to solve practical problems in elegant ways. In particular, he is very interested in 3D shape representations which can enable deep learning on “real-world” geometric datasets which are often noisy, unlabeled, and consisting of very large inputs. He completed my PhD from NYU in 2021 where he worked in the Math and Data Group and the Geometric Computing Lab. His advisors were Joan Bruna and Denis Zorin.
\end{IEEEbiography}

\begin{IEEEbiography}
[{\includegraphics[width=1in,height=1.25in,clip,keepaspectratio]{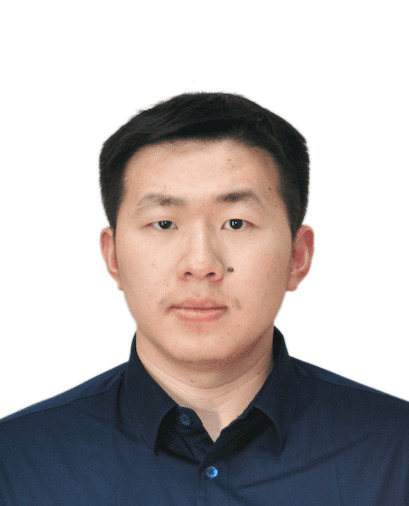}}]{Qun-Ce Xu}
is a postdoctoral researcher in the Department of Computer Science and Technology at Tsinghua University, Beijing, China. He received his Ph.D. degree from the University of Bath, UK, in 2021. His research interests include geometric learning, geometry processing and shape generation.
\end{IEEEbiography}

\begin{IEEEbiography}[{\includegraphics[width=1in,height=1.25in,clip,keepaspectratio]{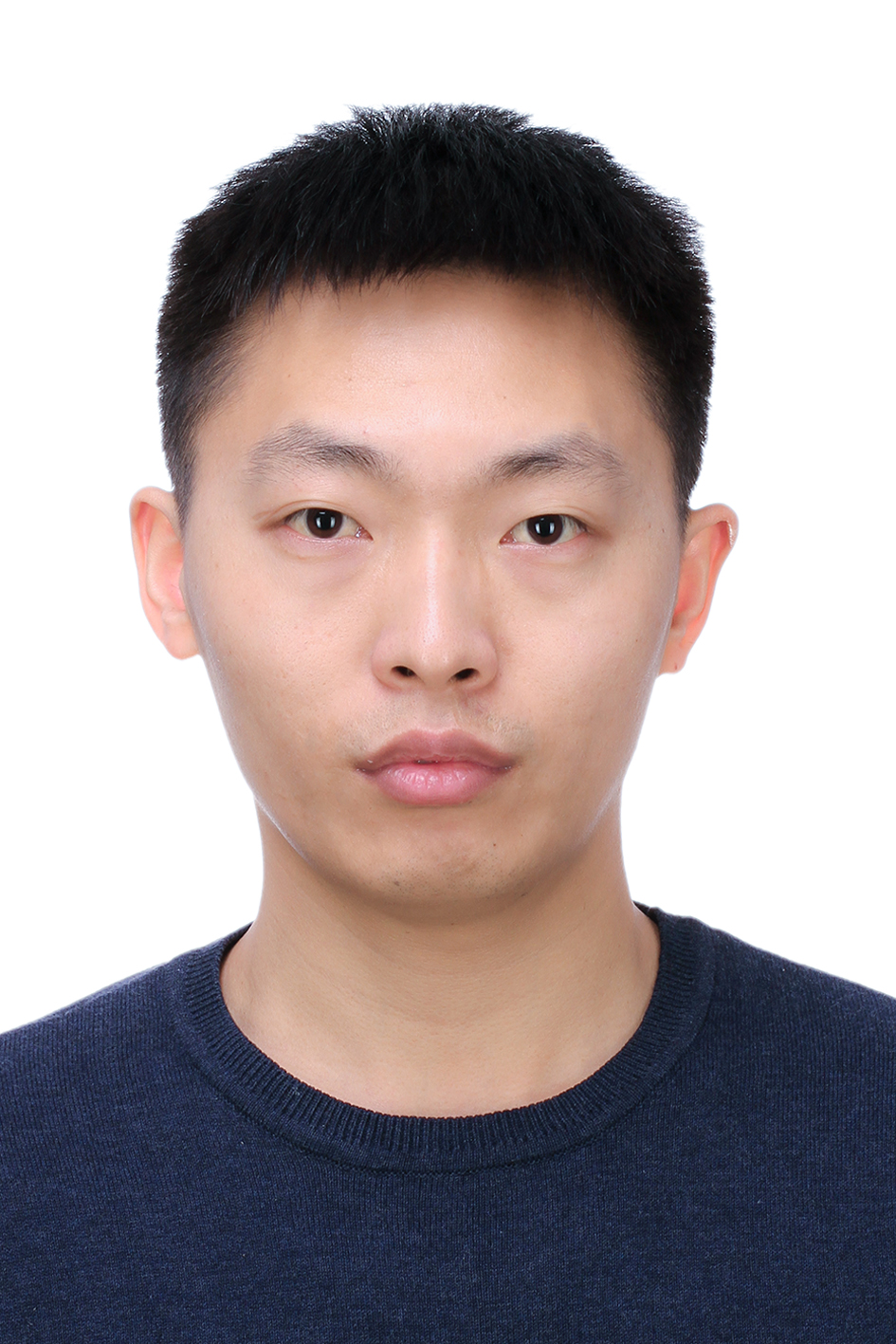}}]{Tai-Jiang Mu} is currently an assistant researcher at Tsinghua University, where he received his B.S. and Ph.D. degrees in Computer Science in 2011 and 2016, respectively. His research interests include computer graphics, computer vision and image processing. He has published more than 40 papers in refereed journals and conferences, including IEEE TPAMI, ACM TOG, IEEE TVCG, IEEE TIP, CVMJ, CVPR, AAAI, ECCV, ICRA, etc. He is the on the editorial board of the Visual Computer.
\end{IEEEbiography}

\begin{IEEEbiography}
[{\includegraphics[width=1in,height=1.25in,clip,keepaspectratio]{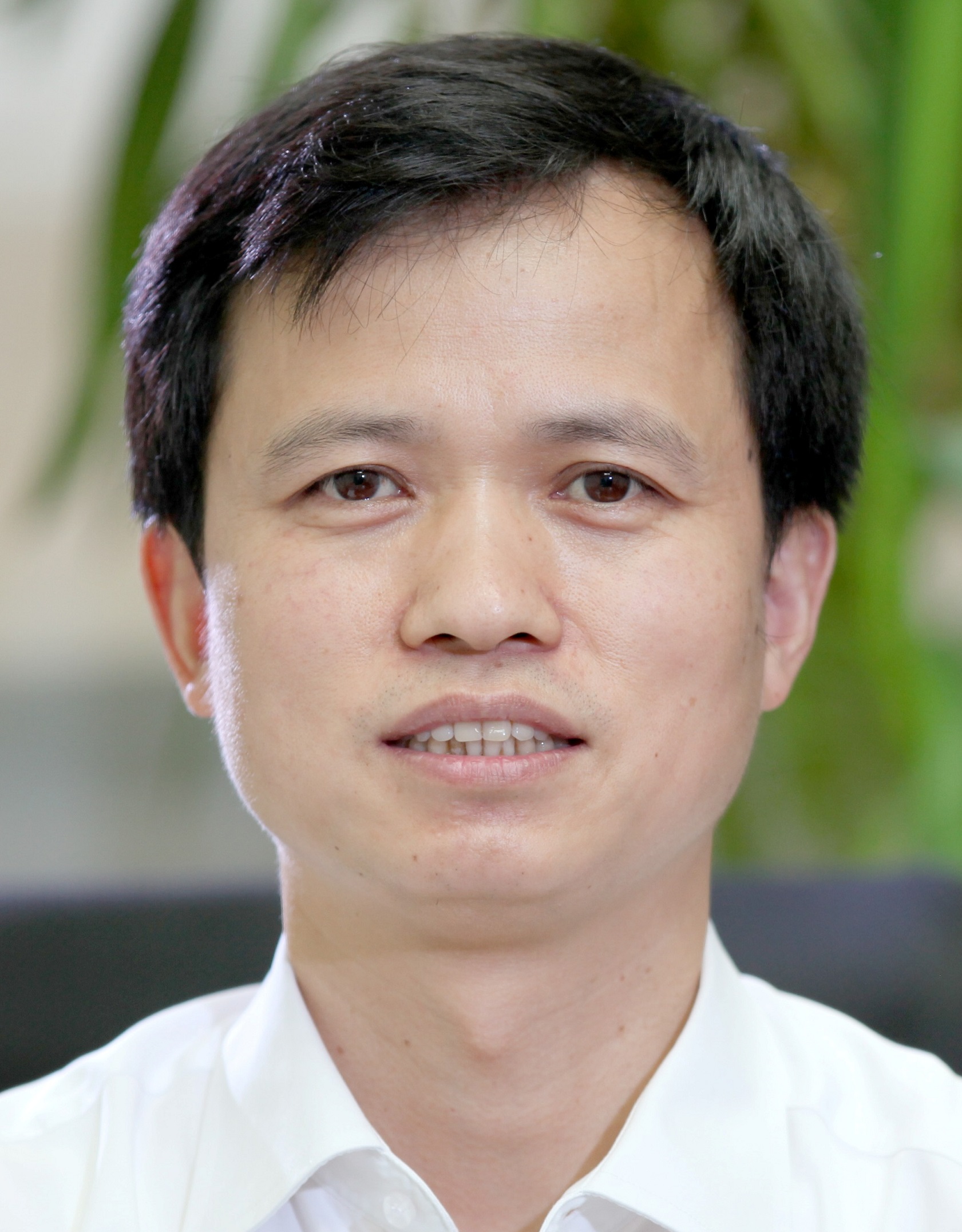}}]{Shi-Min Hu} is currently a professor in Computer Science at Tsinghua University. 
He received a Ph.D. degree from Zhejiang University in 1996. His research interests include geometry processing, image \& video processing, rendering, computer animation, and CAD. 
He has published more than 100 papers in top tier journals and conferences.
He is the Editor-in-Chief of Computational Visual Media, and on the editorial boards of several journals, including Computer Aided Design and Computer \& Graphics.

\end{IEEEbiography}



\vfill
\bibliographystyle{IEEEtran}
\bibliography{references}

@article{tang2022torchsparse,
  title={Torchsparse: Efficient point cloud inference engine},
  author={Tang, Haotian and Liu, Zhijian and Li, Xiuyu and Lin, Yujun and Han, Song},
  journal={Proceedings of Machine Learning and Systems},
  volume={4},
  pages={302--315},
  year={2022}
}

@inproceedings{reddi2018convergence,
  title={On the Convergence of Adam and Beyond},
  author={Reddi, Sashank J and Kale, Satyen and Kumar, Sanjiv},
  booktitle={International Conference on Learning Representations},
  year={2018}
}

@article{changshapenet,
  title={ShapeNet: An Information-Rich 3D Model Repository},
  author={Chang, Angel X and Funkhouser, Thomas and Guibas, Leonidas and Hanrahan, Pat and Huang, Qixing and Li, Zimo and Savarese, Silvio and Savva, Manolis and Song, Shuran and Su, Hao and others}
}

@inproceedings{choy20163d,
  title={3d-r2n2: A unified approach for single and multi-view 3d object reconstruction},
  author={Choy, Christopher B and Xu, Danfei and Gwak, JunYoung and Chen, Kevin and Savarese, Silvio},
  booktitle={Computer Vision--ECCV 2016: 14th European Conference, Amsterdam, The Netherlands, October 11-14, 2016, Proceedings, Part VIII 14},
  pages={628--644},
  year={2016},
  organization={Springer}
}

@inproceedings{mescheder2019occupancy,
  title={Occupancy networks: Learning 3d reconstruction in function space},
  author={Mescheder, Lars and Oechsle, Michael and Niemeyer, Michael and Nowozin, Sebastian and Geiger, Andreas},
  booktitle={Proceedings of the IEEE/CVF conference on computer vision and pattern recognition},
  pages={4460--4470},
  year={2019}
}

@inproceedings{kazhdan2006poisson,
  title={Poisson surface reconstruction},
  author={Kazhdan, Michael and Bolitho, Matthew and Hoppe, Hugues},
  booktitle={Proceedings of the fourth Eurographics symposium on Geometry processing},
  volume={7},
  number={4},
  year={2006}
}

@article{kazhdan2013screened,
  title={Screened poisson surface reconstruction},
  author={Kazhdan, Michael and Hoppe, Hugues},
  journal={ACM Transactions on Graphics (ToG)},
  volume={32},
  number={3},
  pages={1--13},
  year={2013},
  publisher={ACM New York, NY, USA}
}

@inproceedings{peng2020convolutional,
  title={Convolutional occupancy networks},
  author={Peng, Songyou and Niemeyer, Michael and Mescheder, Lars and Pollefeys, Marc and Geiger, Andreas},
  booktitle={Computer Vision--ECCV 2020: 16th European Conference, Glasgow, UK, August 23--28, 2020, Proceedings, Part III 16},
  pages={523--540},
  year={2020},
  organization={Springer}
}

@article{huang2022neural,
  title={A neural galerkin solver for accurate surface reconstruction},
  author={Huang, Jiahui and Chen, Hao-Xiang and Hu, Shi-Min},
  journal={ACM Transactions on Graphics (TOG)},
  volume={41},
  number={6},
  pages={1--16},
  year={2022},
  publisher={ACM New York, NY, USA}
}

@inproceedings{huang2023neural,
  title={Neural kernel surface reconstruction},
  author={Huang, Jiahui and Gojcic, Zan and Atzmon, Matan and Litany, Or and Fidler, Sanja and Williams, Francis},
  booktitle={Proceedings of the IEEE/CVF Conference on Computer Vision and Pattern Recognition},
  pages={4369--4379},
  year={2023}
}

@inproceedings{chang2017matterport3d,
  title={Matterport3D: Learning from RGB-D Data in Indoor Environments},
  author={Chang, Angel and Dai, Angela and Funkhouser, Thomas and Halber, Maciej and Niebner, Matthias and Savva, Manolis and Song, Shuran and Zeng, Andy and Zhang, Yinda},
  booktitle={2017 International Conference on 3D Vision (3DV)},
  pages={667--676},
  year={2017},
  organization={IEEE Computer Society}
}

@inproceedings{calakli2011ssd,
  title={SSD: Smooth signed distance surface reconstruction},
  author={Calakli, Fatih and Taubin, Gabriel},
  booktitle={Computer Graphics Forum},
  volume={30},
  number={7},
  pages={1993--2002},
  year={2011},
  organization={Wiley Online Library}
}

@inproceedings{carr2001reconstruction,
  title={Reconstruction and representation of 3D objects with radial basis functions},
  author={Carr, Jonathan C and Beatson, Richard K and Cherrie, Jon B and Mitchell, Tim J and Fright, W Richard and McCallum, Bruce C and Evans, Tim R},
  booktitle={Proceedings of the 28th annual conference on Computer graphics and interactive techniques},
  pages={67--76},
  year={2001}
}

@inproceedings{williams2019deep,
  title={Deep geometric prior for surface reconstruction},
  author={Williams, Francis and Schneider, Teseo and Silva, Claudio and Zorin, Denis and Bruna, Joan and Panozzo, Daniele},
  booktitle={Proceedings of the IEEE/CVF conference on computer vision and pattern recognition},
  pages={10130--10139},
  year={2019}
}

@inproceedings{curless1996volumetric,
  title={A volumetric method for building complex models from range images},
  author={Curless, Brian and Levoy, Marc},
  booktitle={Proceedings of the 23rd annual conference on Computer graphics and interactive techniques},
  pages={303--312},
  year={1996}
}

@inproceedings{tatarchenko2019single,
  title={What do single-view 3d reconstruction networks learn?},
  author={Tatarchenko, Maxim and Richter, Stephan R and Ranftl, Ren{\'e} and Li, Zhuwen and Koltun, Vladlen and Brox, Thomas},
  booktitle={Proceedings of the IEEE/CVF conference on computer vision and pattern recognition},
  pages={3405--3414},
  year={2019}
}

@incollection{ohtake2005multi,
  title={Multi-level partition of unity implicits},
  author={Ohtake, Yutaka and Belyaev, Alexander and Alexa, Marc and Turk, Greg and Seidel, Hans-Peter},
  booktitle={Acm Siggraph 2005 Courses},
  pages={173--es},
  year={2005}
}

@inproceedings{vizzo2021poisson,
  title={Poisson surface reconstruction for LiDAR odometry and mapping},
  author={Vizzo, Ignacio and Chen, Xieyuanli and Chebrolu, Nived and Behley, Jens and Stachniss, Cyrill},
  booktitle={2021 IEEE international conference on robotics and automation (ICRA)},
  pages={5624--5630},
  year={2021},
  organization={IEEE}
}

@article{chu2021unsupervised,
  title={Unsupervised shape completion via deep prior in the neural tangent kernel perspective},
  author={Chu, Lei and Pan, Hao and Wang, Wenping},
  journal={ACM Transactions on Graphics (TOG)},
  volume={40},
  number={3},
  pages={1--17},
  year={2021},
  publisher={ACM New York, NY}
}

@inproceedings{williams2022neural,
  title={Neural fields as learnable kernels for 3d reconstruction},
  author={Williams, Francis and Gojcic, Zan and Khamis, Sameh and Zorin, Denis and Bruna, Joan and Fidler, Sanja and Litany, Or},
  booktitle={Proceedings of the IEEE/CVF Conference on Computer Vision and Pattern Recognition},
  pages={18500--18510},
  year={2022}
}

@inproceedings{williams2021neural,
  title={Neural splines: Fitting 3d surfaces with infinitely-wide neural networks},
  author={Williams, Francis and Trager, Matthew and Bruna, Joan and Zorin, Denis},
  booktitle={Proceedings of the IEEE/CVF Conference on Computer Vision and Pattern Recognition},
  pages={9949--9958},
  year={2021}
}

@inproceedings{chen2022circle,
  title={Circle: Convolutional implicit reconstruction and completion for large-scale indoor scene},
  author={Chen, Hao-Xiang and Huang, Jiahui and Mu, Tai-Jiang and Hu, Shi-Min},
  booktitle={European Conference on Computer Vision},
  pages={506--522},
  year={2022},
  organization={Springer}
}

@inproceedings{boulch2022poco,
  title={Poco: Point convolution for surface reconstruction},
  author={Boulch, Alexandre and Marlet, Renaud},
  booktitle={Proceedings of the IEEE/CVF Conference on Computer Vision and Pattern Recognition},
  pages={6302--6314},
  year={2022}
}

@inproceedings{ummenhofer2021adaptive,
  title={Adaptive surface reconstruction with multiscale convolutional kernels},
  author={Ummenhofer, Benjamin and Koltun, Vladlen},
  booktitle={Proceedings of the IEEE/CVF International Conference on Computer Vision},
  pages={5651--5660},
  year={2021}
}

@article{wang2022dual,
  title={Dual octree graph networks for learning adaptive volumetric shape representations},
  author={Wang, Peng-Shuai and Liu, Yang and Tong, Xin},
  journal={ACM Transactions on Graphics (TOG)},
  volume={41},
  number={4},
  pages={1--15},
  year={2022},
  publisher={ACM New York, NY, USA}
}

@article{peng2021shape,
  title={Shape as points: A differentiable poisson solver},
  author={Peng, Songyou and Jiang, Chiyu and Liao, Yiyi and Niemeyer, Michael and Pollefeys, Marc and Geiger, Andreas},
  journal={Advances in Neural Information Processing Systems},
  volume={34},
  pages={13032--13044},
  year={2021}
}

@article{hanocka2020point2mesh,
  title={Point2Mesh: a self-prior for deformable meshes},
  author={Hanocka, Rana and Metzer, Gal and Giryes, Raja and Cohen-Or, Daniel},
  journal={ACM Transactions on Graphics (TOG)},
  volume={39},
  number={4},
  pages={126--1},
  year={2020},
  publisher={ACM New York, NY, USA}
}

@inproceedings{park2019deepsdf,
  title={Deepsdf: Learning continuous signed distance functions for shape representation},
  author={Park, Jeong Joon and Florence, Peter and Straub, Julian and Newcombe, Richard and Lovegrove, Steven},
  booktitle={Proceedings of the IEEE/CVF conference on computer vision and pattern recognition},
  pages={165--174},
  year={2019}
}

@inproceedings{atzmonsald,
  title={SALD: Sign Agnostic Learning with Derivatives},
  author={Atzmon, Matan and Lipman, Yaron},
  booktitle={International Conference on Learning Representations}
}

@inproceedings{lipman2021phase,
  title={Phase Transitions, Distance Functions, and Implicit Neural Representations},
  author={Lipman, Yaron},
  booktitle={Proceedings of the 38th International Conference on Machine Learning},
  pages={6702--6712},
  year={2021}
}

@article{wang2017cnn,
  title={O-cnn: Octree-based convolutional neural networks for 3d shape analysis},
  author={Wang, Peng-Shuai and Liu, Yang and Guo, Yu-Xiao and Sun, Chun-Yu and Tong, Xin},
  journal={ACM Transactions On Graphics (TOG)},
  volume={36},
  number={4},
  pages={1--11},
  year={2017},
  publisher={ACM New York, NY, USA}
}

@inproceedings{liao2018deep,
  title={Deep marching cubes: Learning explicit surface representations},
  author={Liao, Yiyi and Donne, Simon and Geiger, Andreas},
  booktitle={Proceedings of the IEEE conference on computer vision and pattern recognition},
  pages={2916--2925},
  year={2018}
}

@inproceedings{gropp2020implicit,
  title={Implicit geometric regularization for learning shapes},
  author={Gropp, Amos and Yariv, Lior and Haim, Niv and Atzmon, Matan and Lipman, Yaron},
  booktitle={Proceedings of the 37th International Conference on Machine Learning},
  pages={3789--3799},
  year={2020}
}

@article{sitzmann2020implicit,
  title={Implicit neural representations with periodic activation functions},
  author={Sitzmann, Vincent and Martel, Julien and Bergman, Alexander and Lindell, David and Wetzstein, Gordon},
  journal={Advances in neural information processing systems},
  volume={33},
  pages={7462--7473},
  year={2020}
}

@article{tancik2020fourier,
  title={Fourier features let networks learn high frequency functions in low dimensional domains},
  author={Tancik, Matthew and Srinivasan, Pratul and Mildenhall, Ben and Fridovich-Keil, Sara and Raghavan, Nithin and Singhal, Utkarsh and Ramamoorthi, Ravi and Barron, Jonathan and Ng, Ren},
  journal={Advances in neural information processing systems},
  volume={33},
  pages={7537--7547},
  year={2020}
}

@article{martel2021acorn,
  title={Acorn: adaptive coordinate networks for neural scene representation},
  author={Martel, Julien NP and Lindell, David B and Lin, Connor Z and Chan, Eric R and Monteiro, Marco and Wetzstein, Gordon},
  journal={ACM Transactions on Graphics (TOG)},
  volume={40},
  number={4},
  pages={1--13},
  year={2021},
  publisher={ACM New York, NY, USA}
}

@inproceedings{takikawa2021neural,
  title={Neural geometric level of detail: Real-time rendering with implicit 3d shapes},
  author={Takikawa, Towaki and Litalien, Joey and Yin, Kangxue and Kreis, Karsten and Loop, Charles and Nowrouzezahrai, Derek and Jacobson, Alec and McGuire, Morgan and Fidler, Sanja},
  booktitle={Proceedings of the IEEE/CVF Conference on Computer Vision and Pattern Recognition},
  pages={11358--11367},
  year={2021}
}

@incollection{lorensen1998marching,
  title={Marching cubes: A high resolution 3D surface construction algorithm},
  author={Lorensen, William E and Cline, Harvey E},
  booktitle={Seminal graphics: pioneering efforts that shaped the field},
  pages={347--353},
  year={1998}
}

@article{zhang1996c,
  title={C-curves: an extension of cubic curves},
  author={Zhang, Jiwen},
  journal={Computer Aided Geometric Design},
  volume={13},
  number={3},
  pages={199--217},
  year={1996},
  publisher={Elsevier}
}

@ARTICLE{9839681,
  author={Yuan, Ganzhangqin and Fu, Qiancheng and Mi, Zhenxing and Luo, Yiming and Tao, Wenbing},
  journal={IEEE Transactions on Visualization and Computer Graphics}, 
  title={SSRNet: Scalable 3D Surface Reconstruction Network}, 
  year={2023},
  volume={29},
  number={12},
  pages={4906-4919},
  doi={10.1109/TVCG.2022.3193406}}

@article{li2022vox,
  title={Vox-surf: Voxel-based implicit surface representation},
  author={Li, Hai and Yang, Xingrui and Zhai, Hongjia and Liu, Yuqian and Bao, Hujun and Zhang, Guofeng},
  journal={IEEE Transactions on Visualization and Computer Graphics},
  volume={30},
  number={3},
  pages={1743--1755},
  year={2022},
  publisher={IEEE}
}

@article{wang2023neural,
  title={Neural-imls: Self-supervised implicit moving least-squares network for surface reconstruction},
  author={Wang, Zixiong and Wang, Pengfei and Wang, Pengshuai and Dong, Qiujie and Gao, Junjie and Chen, Shuangmin and Xin, Shiqing and Tu, Changhe and Wang, Wenping},
  journal={IEEE Transactions on Visualization and Computer Graphics},
  year={2023},
  publisher={IEEE}
}

@inproceedings{erler2020points2surf,
  title={Points2surf learning implicit surfaces from point clouds},
  author={Erler, Philipp and Guerrero, Paul and Ohrhallinger, Stefan and Mitra, Niloy J and Wimmer, Michael},
  booktitle={European conference on computer vision},
  pages={108--124},
  year={2020},
  organization={Springer}
}

@inproceedings{ma2021neural,
  title={Neural-Pull: Learning Signed Distance Function from Point clouds by Learning to Pull Space onto Surface},
  author={Ma, Baorui and Han, Zhizhong and Liu, Yu-Shen and Zwicker, Matthias},
  booktitle={International Conference on Machine Learning},
  pages={7246--7257},
  year={2021},
  organization={PMLR}
}

@inproceedings{chabra2020deep,
  title={Deep local shapes: Learning local sdf priors for detailed 3d reconstruction},
  author={Chabra, Rohan and Lenssen, Jan E and Ilg, Eddy and Schmidt, Tanner and Straub, Julian and Lovegrove, Steven and Newcombe, Richard},
  booktitle={European Conference on Computer Vision},
  pages={608--625},
  year={2020},
  organization={Springer}
}

@article{ma2023learning,
  title={Learning signed distance functions from noisy 3d point clouds via noise to noise mapping},
  author={Ma, Baorui and Liu, Yu-Shen and Han, Zhizhong},
  year={2023}
}

@inproceedings{wang2023alto,
  title={Alto: Alternating latent topologies for implicit 3d reconstruction},
  author={Wang, Zhen and Zhou, Shijie and Park, Jeong Joon and Paschalidou, Despoina and You, Suya and Wetzstein, Gordon and Guibas, Leonidas and Kadambi, Achuta},
  booktitle={Proceedings of the IEEE/CVF Conference on Computer Vision and Pattern Recognition},
  pages={259--270},
  year={2023}
}

@inproceedings{li2024implicit,
  title={Implicit filtering for learning neural signed distance functions from 3d point clouds},
  author={Li, Shengtao and Gao, Ge and Liu, Yudong and Gu, Ming and Liu, Yu-Shen},
  booktitle={European Conference on Computer Vision},
  pages={234--251},
  year={2024},
  organization={Springer}
}

@inproceedings{riegler2017octnet,
  title={Octnet: Learning deep 3d representations at high resolutions},
  author={Riegler, Gernot and Osman Ulusoy, Ali and Geiger, Andreas},
  booktitle={Proceedings of the IEEE conference on computer vision and pattern recognition},
  pages={3577--3586},
  year={2017}
}

@inproceedings{ben2022digs,
  title={Digs: Divergence guided shape implicit neural representation for unoriented point clouds},
  author={Ben-Shabat, Yizhak and Koneputugodage, Chamin Hewa and Gould, Stephen},
  booktitle={Proceedings of the IEEE/CVF conference on computer vision and pattern recognition},
  pages={19323--19332},
  year={2022}
}

@article{muller2022instant,
  title={Instant neural graphics primitives with a multiresolution hash encoding},
  author={M{\"u}ller, Thomas and Evans, Alex and Schied, Christoph and Keller, Alexander},
  journal={ACM transactions on graphics (TOG)},
  volume={41},
  number={4},
  pages={1--15},
  year={2022},
  publisher={ACM New York, NY, USA}
}

@inproceedings{koneputugodage2023octree,
  title={Octree guided unoriented surface reconstruction},
  author={Koneputugodage, Chamin Hewa and Ben-Shabat, Yizhak and Gould, Stephen},
  booktitle={Proceedings of the IEEE/CVF Conference on Computer Vision and Pattern Recognition},
  pages={16717--16726},
  year={2023}
}

@inproceedings{lindell2022bacon,
  title={Bacon: Band-limited coordinate networks for multiscale scene representation},
  author={Lindell, David B and Van Veen, Dave and Park, Jeong Joon and Wetzstein, Gordon},
  booktitle={Proceedings of the IEEE/CVF conference on computer vision and pattern recognition},
  pages={16252--16262},
  year={2022}
}
\clearpage\newpage

\appendix
\subsection{Optimal solution to the surface fitting variational problem}
\label{subsec:optimal_surface_fitting}

Here, we provide a detailed derivation that shows how the energy expression in Equation~\ref{eq:energy} is transformed into the closed-form solution presented in Equation~\ref{eq:closed-form}. This transformation is a critical step, as the resulting closed-form expression allows for a direct and efficient computation of the solution.

\begin{equation}
  \begin{aligned}
    E(f) = &\iiint_{\Omega} \left\|\nabla f - \vec{\mathbf{N}}\right\|_2^2 \, \mathrm{d}V \\
           &+ \lambda_1 \iiint_{\Omega} \left\|H(f)\right\|_F^2 \, \mathrm{d}V \\
           &+ \lambda_2 \sum_{p \in \mathcal{P}} \left(f(p) - \xi(p)\right)^2
  \end{aligned}
\end{equation}

where $ 1(\cdot)$  denotes the Dirac delta function, which satisfies $\iiint_\Omega 1(\mathbf{x}) \, d\mathbf{x} = 1$  and  $1(\mathbf{x}) = 0$  for all $\mathbf{x} \neq 0$.

Since $ \left\|H(f)\right\|_F^2$ can be expressed as
\begin{equation}
  \begin{aligned}
    \left\|H(f)\right\|_F^2 = \sum_i\sum_j \left  (\frac{\partial^2 f}{\partial x_i \partial x_j}\right)^2
    \end{aligned}
\end{equation}
The Euler-Lagrange equation is a second-order partial differential equation that describes the motion of a system of conservative forces. In the context of surface reconstruction, it is used to find the optimal surface fitting by minimizing the energy of the implicit field $f$. By extending the Euler-Lagrange equation into a multi-dimensional context, we identify that the stationary points of the energy function $E$ conform to the conditions:

\begin{equation}
  \resizebox{\linewidth}{!}{
    \begin{minipage}{\linewidth}
      \begin{align*}
        \medmuskip=0.5mu 
        \thinmuskip=0.5mu 
        \thickmuskip=0.5mu 
        &\Delta f-\nabla\cdot\vec{\mathbf{N}}+\lambda_1\sum_i\sum_j\frac{\partial^2}{\partial x_i\partial x_j}(\frac{\partial^2 f}{\partial x_i \partial x_j})\\
        &+\lambda_2\sum_{p\in\mathcal{P}}(f(p)-\xi(p))\mathbb{1}(x-p)=0\quad i,j\in\{1,2,3\}
        \tag{\theequation}
      \end{align*}
      \label{eq:EL}
    \end{minipage}
  }
\end{equation}

The third term of Equation~\ref{eq:EL} can be simplified as follows:
\begin{equation}
  \resizebox{\linewidth}{!}{
    \begin{minipage}{\linewidth}
      \begin{align*}
        \sum_i\sum_j\frac{\partial^2}{\partial x_i\partial x_j}(\frac{\partial^2 f}{\partial x_i \partial x_j})
        =\sum_i\sum_j\frac{\partial^2}{\partial x_i\partial x_i}(\frac{\partial^2 f}{\partial x_j \partial x_j})
        =\Delta(\Delta f)\quad \\i,j\in\{1,2,3\}
        \tag{\theequation}
      \end{align*}
    \end{minipage}
  }
\end{equation}

This simplification allows us to represent the original Equation~\ref{eq:EL} in a more concise form:
\begin{equation}
  \resizebox{\linewidth}{!}{
    \begin{minipage}{\linewidth}
      \begin{align*}
        \Delta f-\nabla\cdot\vec{\mathbf{N}}+\lambda_1\Delta(\Delta f)
        +\lambda_2\sum_{p\in\mathcal{P}}(f(p)-\xi(p))\mathbb{1}(x-p)=0
        \tag{\theequation}
      \end{align*}
    \end{minipage}
  }
\end{equation}

Given that a differentiable functional is stationary at its local extrema, the quasi-convex nature of the problem almost certainly yields an optimal solution to the surface fitting energy, which is addressed by our novel geometric solver .

Following the Galerkin method, we transform the strong formulation in Equation (15) into its weak form by introducing a test function \(\omega\), leading to: 
\begin{equation}
  \resizebox{\linewidth}{!}{
    \begin{minipage}{\linewidth}
      \begin{align*}
        \forall\omega,\iiint_{\Omega}\omega\left(\Delta f-\nabla\cdot\vec{\mathbf{N}}+\lambda_1\Delta(\Delta f)\right)\mathrm{d}x&\\
        +\lambda_2\sum_{p\in\mathcal{P}}\omega(p)\left(f(p)-\xi(p)\right)&=0
        \tag{\theequation}
      \end{align*}
    \end{minipage}}
\end{equation}
The first integral term can be simplified by the integration by parts as 
$\omega \Delta f = \nabla \cdot (\omega \nabla f) - \nabla f \cdot \nabla \omega,$
and by applying Stokes' theorem, which incorporates the boundary conditions over the domain \(\Omega\). Subsequently, we replace the arbitrary test functions with the predetermined basis functions, where 
$f(\mathbf{p}) = \sum_k \bar{\alpha}_k \bar{B}_k(\mathbf{p}),$
using the product subscript notation, resulting in the following discretization: 

\begin{equation}
  \resizebox{\linewidth}{!}{
    \begin{minipage}{\linewidth}
      \footnotesize
      \medmuskip=0.5mu
      \thinmuskip=0.5mu
      \thickmuskip=0.5mu
      \begin{align*}
        \forall\bar{k},\quad & \sum_{\bar{l}}\alpha_{\bar{l}}\iiint_{\Omega}\nabla\mathcal{B}_{\bar{k}}^{\top}\nabla\mathcal{B}_{\bar{l}}\mathrm{d}V+\lambda_2\sum_{\bar{l}}\alpha_{\bar{l}}\iiint_{\Omega}\Delta\mathcal{B}_{\bar{k}}\cdot\Delta \mathcal{B}_{\bar{l}}\mathrm{d}V \\
        &+\lambda_1\sum_{\bar{l}}\alpha_{\bar{l}}\sum_{p\in\mathcal{P}}\mathcal{B}_{\bar{k}}(\boldsymbol{p})\mathcal{B}_{\bar{l}}(\boldsymbol{p}) \\
        &= \iiint_{\Omega}\nabla\mathcal{B}_{\bar{k}}^{\top}\vec{\mathbf{N}}\mathrm{d}V+\lambda_1\sum_{p\in\mathcal{P}}\mathcal{B}_{\bar{k}}(\boldsymbol{p})\xi(\boldsymbol{p})
        \tag{\theequation}
      \end{align*}
    \end{minipage}}
\end{equation}

which perfectly aligns with the form 
$L\boldsymbol{\alpha} = \mathbf{d}$
as specified in Equation~\ref{eq:energy}.


\subsection{Basis functions}
\label{subsec:basis_function}
The selection of the basis functions $B(\boldsymbol{p})$, as introduced in Equation~\ref{eq:basis}, is a critical design choice. An appropriate parameterization must be chosen carefully, as it needs to strike a balance between two competing objectives: representational power and computational efficiency. The basis must be diverse enough to represent a wide range of functions, yet simple enough to ensure efficient runtime performance, particularly for the integration required in Equation~\ref{eq:closed-form}. To address this, we employ a parameterization specifically designed to optimize this trade-off.

We choose the set of elementary functions to be power functions, such as $q_u \in \{1, x, x^2, x^3, \ldots\}$, which make $b(x)$ a polynomial. The piecewise definition of $B(\boldsymbol{p})$ further leads us to the constraint that $b(-1.5) = b(1.5) = 0$ and its derivatives $\frac{\partial b}{\partial x}(\pm1.5) = 0$ for a $C^1$-continuous implicit field. To ensure this constraint, we set the coefficients $\boldsymbol{m}_{k}^{s}$ to lie within a linear null space defined by these constraints. Although the chosen family of bases consists of simple functions, it still maintains the ability to represent highly complex shapes, similar to the concept of Taylor series. Furthermore, to incorporate high-frequency biases, we introduce sine waves to the set of elementary functions. This Fourier-style family is widely adopted in traditional geometric modeling \cite{zhang1996c} and modern deep learning \cite{tancik2020fourier}. Although this degrades the $C^1$-continuity to $C^0$, it empirically helps with training.

\end{document}